\documentclass{ws-ijprai}
\usepackage{xcolor}
\usepackage[verbose,hypertexnames=false]{hyperref}
\hypersetup{colorlinks=false,allbordercolors=blue,pdfborderstyle={/S/U/W 1}}
\usepackage{xcolor,soul,framed} %,caption

\colorlet{shadecolor}{yellow}
\DeclareGraphicsExtensions{.pdf,.jpeg,.png}

\usepackage{array}
\usepackage{mdwmath}
\usepackage{mdwtab}
\usepackage{eqparbox}
\usepackage{url}

\usepackage{algorithm}
\usepackage{algorithmic}

\usepackage{amsfonts}
\usepackage{graphicx}
\usepackage{tabularx}
\usepackage{bm}
\usepackage{multirow}
\usepackage{subfigure}
\usepackage{booktabs}

\begin{document}

\markboth{Q.W. Wang, Y. Song and S.T. Xia}{Bridging Weakly-Supervised Learning and VLM Distillation}

%%%%%%%%%%%%%%%%%%%%% Publisher's area please ignore %%%%%%%%%%%%%%%
%
\catchline{}{}{}{}{}
%
%%%%%%%%%%%%%%%%%%%%%%%%%%%%%%%%%%%%%%%%%%%%%%%%%%%%%%%%%%%%%%%%%%%%

\title{Bridging Weakly-Supervised Learning and VLM Distillation: Noisy Partial Label Learning for Efficient Downstream Adaptation}

\author{Qian-Wei~Wang$^{\dagger,\ddagger}$, Yaguang~Song$^{\dagger,}$\footnote{Corresponding author.}~ and Shu-Tao~Xia$^{\ddagger,\dagger}$}

\address{
$^{\dagger}$
Institute of Perceptual Intelligence, Peng Cheng Laboratory \\
Shenzhen, Guangdong 518055, China
}

\address{
$^{\ddagger}$
Tsinghua Shenzhen International Graduate School, Tsinghua University \\
Shenzhen, Guangdong 518055, China \\
\email{wanggw21@mails.tsinghua.edu.cn, songyg01@pcl.ac.cn, xiast@sz.tsinghua.edu.cn}
}

\maketitle

% \begin{history}
% \received{(Day Month Year)}
% \revised{(Day Month Year)}
% \accepted{(Day Month Year)}
% \published{(Day Month Year)}
% %\comby{(xxxxxxxxxx)}
% \end{history}

\begin{abstract}
In the context of noisy partial label learning (NPLL), each training sample is associated with a set of candidate labels annotated by multiple noisy annotators. With the emergence of high-performance pre-trained vision-language models (VLMs) such as CLIP, LLaVA, and GPT-4V, leveraging these models to replace time-consuming manual annotation and enable annotation-free training has become a promising research direction. This paper studies learning from noisy partial labels generated by pre-trained VLMs and proposes a collaborative consistency regularization (Co-Reg) framework. Unlike symmetric noise commonly assumed in traditional noisy label learning, VLM-generated noise is instance-dependent and reflects the intrinsic biases of pre-trained models, posing greater challenges. To address this issue, we jointly train two neural networks to perform collaborative label purification via a co-pseudo-labeling mechanism, while enforcing consistency regularization in both label and feature representation spaces. In addition, multiple anti-overfitting strategies are introduced, including alternating optimization of contrastive representations and pseudo-labels, as well as maintaining class prototypes in a shared feature space. The proposed method can further incorporate few-shot manually annotated labels for performance enhancement. Extensive experiments under various settings demonstrate the effectiveness of our approach and highlight the potential of integrating weakly supervised learning into the knowledge distillation of pre-trained models.
\end{abstract}


\section{Introduction}

Partial label learning (PLL) \cite{DBLP:conf/icml/LvXF0GS20,DBLP:conf/ijcai/ZhangY15a,DBLP:journals/tkde/LyuFWLL21,fan2021partial,fan2022partial,lu2025partial}, a key branch of weakly-supervised learning, addresses the classification problem where each training sample is associated with multiple candidate labels, with only one being the ground-truth. Due to the core assumption that the ground-truth label must lie within the candidate label set often failing to hold in practical applications, noisy partial label learning (NPLL) \cite{qiao2023fredis,shi2023unreliable,shi2023robust,wang2022pico+} has garnered increasing attention in recent years. This scenario relaxes the constraints of PLL by allowing the ground-truth labels of some noisy samples to exist outside their candidate label sets.

Most existing NPLL studies rely on partially labeled data from manual annotation. With the advent of the pre-train large model era, researchers have enabled multi-modal models with massive parameters to learn from massive ``image-text" pairs in general domains, aligning their visual and textual encoders into a unified space and leveraging the models' ability to understand textual instructions for generalization to unseen tasks. This naturally gives rise to a research direction in weakly-supervised learning scenarios like NPLL: using these models to replace tedious manual annotation for automatic training sample labeling and downstream task-specific model training.

This paper constructs a pipeline that uses pre-train VLM to annotate noisy candidate label sets for downstream task images and then trains a specialized model based on noisy partial label algorithms. Specifically, for a VLM annotator, each prompt template yields a classification result, and the results from multiple templates collectively form the candidate label sets. Unlike the randomly constructed symmetric noise label matrices commonly used in traditional NPLL research, the noise annotated by VLMs in this study is instance-dependent, exhibiting underlying patterns influenced by the knowledge embedded in pre-train models. This significantly increases model training difficulty since that random noise is easily identified as conflicting labels by algorithms due to its lack of latent regularity and can be corrected via pseudo-labeling, while noise generated by pre-train models is prone to being misjudged as high-confidence true labels because it aligns with the patterns distilled from the teacher pre-train model.

To address this, we propose a novel \textbf{Co}llaborative consistency \textbf{Reg}ularization (Co-Reg) method. This method simultaneously trains two neural networks to achieve collaborative purification of training labels through a co-pseudo-labeling mechanism, while enforcing consistency regularization constraints in both label space and feature representation space. Specifically, the two neural networks separately partition the partially labeled samples from pre-train models into reliable and noisy subsets, which are then provided to the other network for training. This alleviates the confirmation bias caused by mimicking pre-train models, i.e., the increasing predicted confidence in noisy labels during iteration. For samples detected with noisy partial labels, the algorithm treats them as unlabeled samples and aggregates the prediction outputs of data-augmented variants from both networks to obtain an optimized class distribution. This effective utilization rather than discarding of noisy samples corrects the annotation errors of pre-train models on downstream tasks and significantly enhances the performances of specialized models compared to the zero-shot generalization of original pre-train models. During self-training, the model not only uses the generated pseudo-labels as the optimization target for class distribution but also maintains a representation prototype for each class in the shared projected embedding space of both networks. By calculating the similarity distribution between the current sample representation and each class's representation prototype and aligning it with the sample's pseudo-label distribution, unified calibration is achieved. Additionally, our method introduces a noise-tolerant supervised contrastive learning module to enhance the model's ability to learn discriminative feature representations for specified downstream tasks.

In our experiments, we compare our method with state-of-the-art noisy single/partial label learning algorithms. To ensure fairness, both the noisy single-labels and noisy partial labels used here are generated by the same pre-trained VLM as the annotator. This study discusses the advantages and disadvantages of incorporating NPLL into knowledge distillation for pre-trained VLMs on specified downstream tasks. While fully fine-tuning typically yields the largest performance gains, it requires extensive expensive manual annotation of downstream task samples, which is infeasible in many scenarios. Although few-shot fine-tuning techniques such as prompt learning, adapter, and LoRA demand fewer labeled samples, they still rely on a small amount of manually annotated data. Additionally, since these methods attach minimal trainable parameters to pre-trained models, they cannot produce inference-efficient models tailored to downstream tasks. Compared with traditional unsupervised knowledge distillation, NPLL algorithms achieve significant performance improvements through strategies like consistency regularization. Comparing experiments are conducted for knowledge distillation and few-shot fine-tuning to demonstrate the feasibility of integrating NPLL into pre-trained model distillation. Furthermore, we extend our method to few-shot scenarios, enabling it to leverage a small number of manually annotated real-world labels to further enhance performance, thus improving the method’s real-world applicability.

Our main contributions can be summarized as:
\begin{itemize}
    \item We investigate the NPLL framework formalized from annotation from pre-trained VLMs. To address the instance-dependent noise that inherits the latent patterns of these pre-trained models, we propose a novel collaborative consistency regularization approach.

    \item We employ multiple VLMs to annotate image datasets from diverse scenarios and conduct experiments on these datasets. We compare the results of our method with state-of-the-art weakly supervised learning algorithms under noisy single/partial label settings.
    
    \item Additionally, we compare our method with widely used pre-trained model application paradigms such as knowledge distillation and few-shot fine-tuning, demonstrating the great potential of incorporating NPLL into pre-trained model distillation. This inspiration can be further extended to various types of downstream tasks, pre-trained models and weakly-supervised problems.

    \item Our method is also extended to few-shot learning scenarios, enabling it to leverage a small number of manually annotated valid labels to further improve performance.
    
\end{itemize}

\section{Related Work}
\subsection{Weakly-supervised learning}
Weakly-supervised learning \cite{zhou2018brief,wang2020learning,zhang2018learning} research related to this work mainly originates from two directions: learning from noisy labels and learning from candidate label sets (i.e. partial labels). This section primarily introduces the research works of the above two directions and their intersection: noisy partial label learning (NPLL).

Learning with noisy labels is a critical subfield of weakly-supervised learning, addressing scenarios where training data contains erroneous labels due to human errors, noisy annotations, or ambiguous data collection processes. Early works, such as \cite{reed2014training}, introduced bootstrapping to handle noisy labels by leveraging perceptual consistency, allowing models to identify reliable samples through feature similarity. Later, \cite{han2018co-teaching} proposed co-teaching, a dual-network framework where two models iteratively select clean samples for mutual training, demonstrating robustness to extreme noise rates. These methods capitalized on the observation that deep networks memorize clean data before noisy samples.

Recent advancements focus on loss function design and noise modeling. \cite{zhou2021asymmetric} introduced asymmetric loss functions, which adaptively penalize misclassifications based on noise type, outperforming symmetric alternatives. \cite{kim2024learning} proposed an EM-based framework (LNL-Flywheel) that integrates two expectation-maximization cycles to distinguish clean labels and refurbish corrupted ones, achieving state-of-the-art results on benchmarks like CIFAR-10/100. Theoretical analyses, such as \cite{xu2021theoretical}, revealed two training phases: clean data prioritization followed by noise memorization, explaining the efficacy of early stopping and sample selection.

Current research increasingly integrates self-supervised learning to enhance noise robustness and combines with large-scale pre-trained models, achieving further breakthroughs in real-world applications. \cite{chen2024learning} studied noise in pre-trained foundation models, showing that even slight pre-training noise degrades out-of-domain generalization. They proposed NMTune, a tuning method which is applicable in both parameter-efficient and black-box manners, to affine the feature space to mitigate noise effects. \cite{zhang2023noisy} addressed long-tailed noisy data with RCAL, a representation calibration approach combining contrastive learning and Gaussian distribution modeling to handle class imbalance and label corruption. Despite progress, challenges remain, including asymmetric noise (e.g., class/instance-dependent corruption) and integrating denoising into a more automated pipeline for downstream tasks.

Partial label learning (PLL) \cite{feng2018leveraging,wang2019partial,wang2023deep} aims to solve the classification problem where each training sample is associated with a set of candidate labels, with exactly one being the ground-truth label. Its core challenge lies in disambiguating the ground-truth label from other false-positive candidate labels. Early research in this field can be traced back to the pioneering works like \cite{Hullermeier_Beringer2006} and \cite{Nguyen_Caruana2008}. Cour et al. \cite{Cour_2011} first systematically defined the PLL problem and proposed a convex optimization-based learning framework, laying the theoretical foundation for subsequent research. Building on previous work, Zhang and Yang \cite{DBLP:conf/ijcai/ZhangY15a} optimized the label disambiguation process through instance-level methods, propelling the practical development of PLL. 

With the prevailing of deep learning, PLL methods have shifted toward data-augmentation-based consistency regularization, leading to the emergence of a series of methods with impressive results even be on par with their fully-supervised counterparts. Wang et al. \cite{wang2022pico} first employed contrastive learning in PLL algorithms, using an iterative approach of representation optimization and class distribution optimization to disambiguate candidate label sets. Theoretically, they formalized the algorithm as a variant of the expectation-maximization (EM) algorithm. Wu et al. \cite{wu2022revisiting} revisited consistency regularization and proposed a novel framework that explicitly models the uncertainty in partial labels. Their method leverages multiple augmented views of each sample to enforce prediction consistency, while simultaneously minimizing the probability of non-candidate labels.

However, traditional PLL imposes an overly strict assumption that the ground-truth labels of all samples must be included within the their candidate label sets, which can hardly be met in scenarios such as crowd-sourcing and learning with pre-trained model annotated partial labels. As a result, there has been a growing tendency to study a more practical extension of PLL, known as noisy partial label learning (NPLL) \cite{lian2022arnet,wang2022pico+,xu2024alim,shi2023unreliable,shi2023robust,qiao2023fredis}. NPLL allows the existence of noises of partial labels, i.e. the ground-truth label is not any of the candidate labels. Previous methods usually design specific mechanisms to handle noisy partial samples simultaneously with partial label disambiguation. Some methods \cite{qiao2023fredis,lian2022arnet,shi2023robust} incorporate non-candidate labels with high predicted probabilities into the candidate label sets during training. Shi et al. \cite{shi2023unreliable} divide the samples based on whether they contain noise. Xu et al. \cite{xu2024alim} assign a certain probability to non-candidate labels in the training objective.

Despite these achievements, previous NPLL researches have primarily addressed simulated settings such as random noise and lack experimentation in more challenging scenarios, such as learning from partial labels in real-world annotations. Some papers \cite{xu2021instance,xu2023progressive,wu2024distilling} investigated instance-dependent partial labels by training a neural network on ground-truth labels to model the probability of false-positive candidate labels being flipped. While this approach significantly advances PLL algorithms toward real-world applicability, it remains constrained to simulated scenarios. Moreover, the study does not account for the scenario where partial labels inherently contain noise, limiting its generalizability to more complex, noisy annotation environments.

\begin{figure*}
    \centering
    \includegraphics[width=0.9\linewidth]{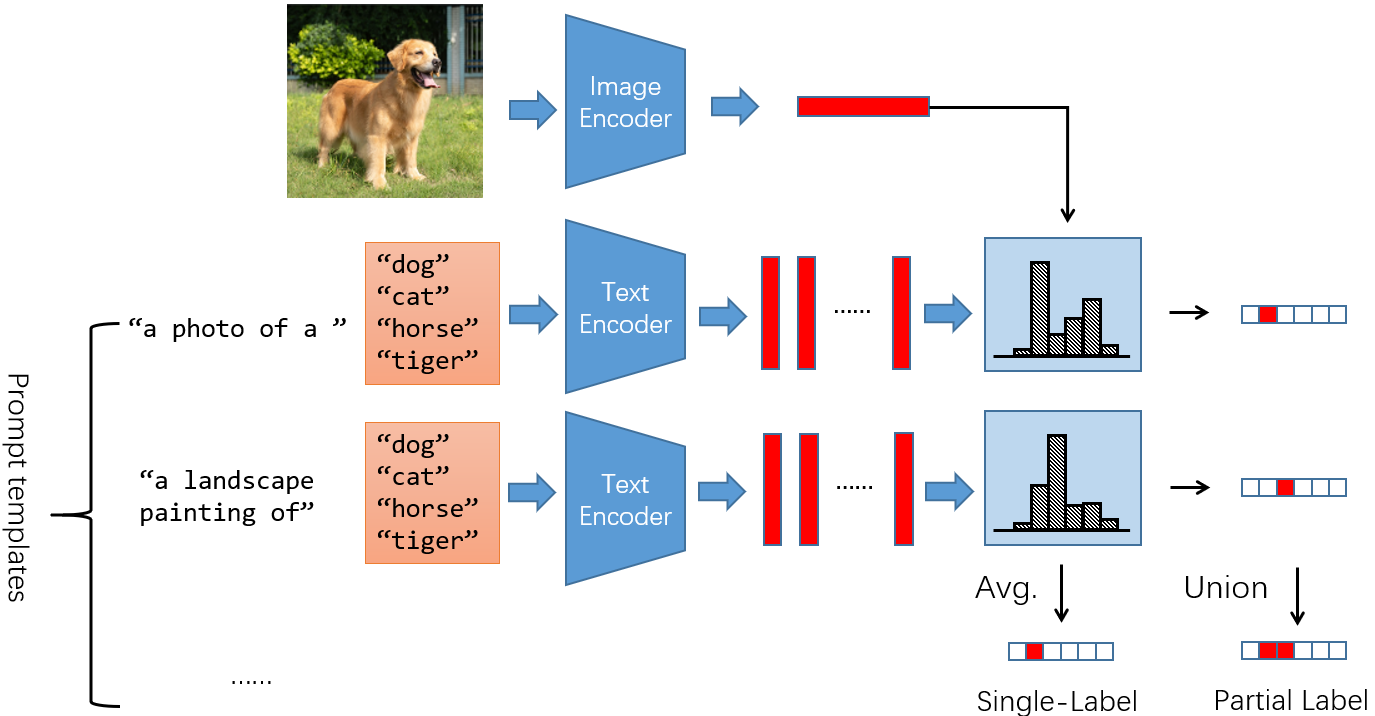}
    \caption{Schematic diagram of using CLIP and multiple prompt templates to annotate images from downstream tasks with noisy partial labels (candidate label sets). In this process, each prompt template is combined with all class names of the task to form text inputs, which are then encoded by the text encoder to obtain text embeddings. These text embeddings are matched with image embeddings (derived from the image encoder) to generate CLIP's predicted class distribution for each image.}
    \label{clip_annotation}
\end{figure*}
\subsection{Applying pre-trained VLMs to downstream tasks}

Pre-trained vision-language models (VLMs), such as CLIP \cite{radford2021learning}, LLaVA \cite{liu2023visual}, and GPT-4V \cite{openai2024gpt4v}, have revolutionized multi-modal task solving by learning aligned image-text representations from massive datasets. These models exhibit remarkable zero-shot generalization capabilities, enabling direct inference on unseen tasks by matching input images with natural language descriptions of target categories. For example, CLIP can classify images into arbitrary categories defined by text prompts (e.g., ``a photo of a airplane.") without task-specific training, while LLaVA leverages visual grounding and large language model (LLM) reasoning to handle complex visual question-answering tasks in zero-shot settings. However, direct zero-shot application often struggles with \emph{domain shift} (e.g., medical images vs. general-domain training data) and \emph{computational inefficiency} due to large model sizes during inference. Full fine-tuning is the most straightforward approach; however, it requires a large number of labeled samples from downstream tasks, without which it is prone to overfitting. To address these challenges, researchers have developed lightweight adaptation techniques.

\subsubsection{Prompt learning for semantic alignment}
Prompt learning aims to bridge the gap between pre-trained VLMs and downstream tasks by optimizing text or visual prompts. For CLIP-like models, text prompt engineering involves designing natural language templates (e.g., ``a satellite remote sensing image of \textless CLASS\_NAME\textgreater.'') to enhance category description specificity. Zhou et al. \cite{zhou2022conditional} proposed conditional prompt tuning, which learns task-specific prompt embeddings while keeping the CLIP encoder frozen. For multi-modal models like LLaVA, visual prompt tuning \cite{jia2022visual} extends this idea by adding learnable visual tokens to the image input, enabling better alignment with LLM-generated responses. For instance, in medical image classification, visual prompts can encode domain-specific features (e.g., X-ray contrast patterns) to improve LLaVA’s diagnostic accuracy on unseen medical datasets.

\subsubsection{Parameter-efficient adaptation: adapters and LoRA}
Adapter-based methods \cite{houlsby2019parameter} and Low-Rank Adaptation (LoRA) \cite{hu2022lora} offer lightweight fine-tuning alternatives. Adapters insert small task-specific modules into the VLM architecture (e.g., between CLIP’s image and text encoders), allowing the model to learn task dynamics without altering pre-trained weights. Gao et al. \cite{gao2024clip} demonstrated that CLIP adapters can achieve 90\% of full fine-tuning performance on CIFAR-10 with only 0.1\% additional parameters. LoRA decomposes weight updates into low-rank matrices, drastically reducing memory usage. For LLaVA, LoRA has been applied to fine-tune the cross-attention layers between the vision encoder and LLM, enabling efficient adaptation to robotics instruction following tasks while preserving general multi-modal reasoning abilities.

\subsubsection{Knowledge distillation for model compression}
Knowledge distillation \cite{hinton2015distilling} transfers knowledge from large VLMs to compact student models. For LLaVA-like models, multi-modal distillation involves aligning both visual features and LLM-generated responses. For example, a student model can be trained to replicate LLaVA’s output distributions on paired image-question-answer datasets, enabling deployment on edge devices with limited computational resources.

\section{Pre-trained VLMs Annotated Partial Labels}\label{annotation}
This section introduces how we use pre-trained VLMs to annotate noisy partial labels for downstream image datasets. Next, we first take CLIP as an example (See Fig. \ref{clip_annotation}). We use a collection of prompt templates, denoted as $\{\mathcal{T}_1(\cdot), \mathcal{T}_2(\cdot) \dots \mathcal{T}_d(\cdot)\}$, and combine them with the class names of downstream task to form the textual input. The template here is like: ``a photo of a \textless CLASS\_NAME\textgreater.". We denote the class names of the classification task as $\{n_1, n_2, \dots, n_C\}$, where $C$ is the number of categories, and the combination of the $i$-th template with $j$-th class can be written as $\mathcal{T}_i(n_j)$.

CLIP is a dual-encoder architecture composed of a image encoder and a text encoder, denoted as \textrm{ImageEncoder} and \textrm{TextEncoder} respectively, which can compute semantic relevance scores between input images and text. For each template $\mathcal{T}_i(\cdot)$, we combine it with all class names of the targeted classification task and then take them as the input of the text encoder of CLIP and obtain $C$ textual representations $\{t_1, t_2, \dots, t_C\}$, where $t_j = \textrm{TextEncoder}(\mathcal{T}_i(n_j))$. Meanwhile, we obtain the image representation by inputting the training image into the image encoder of CLIP, denoted as $r = \textrm{ImageEncoder}(\mathcal{I})$. Then, we can predict the probabilities of the image belonging to different classes under this prompt template as $p_i = \textrm{softmax}(r t_1, r t_2, \dots, r t_C)$. Thus, we can obtain the predicted label of this prompt template by $\hat{p}_i = \arg\max p_i$.

Then, we deem each prompt template's predicted label as a candidate label, which means that it could possibly be the ground-truth label of the sample, and take the union of all candidates to obtain the candidate label set $S$. The partial label we used in the algorithm is formalized as $y = (y_1, y_2, \dots, y_C) \in \{0, 1\}^C$, in which $y_j = 1$ if $j \in S$ and $y_j = 0$ if $j \notin S$. 

In experiments, we also compare the methods that annotate noisy single-labels by averaging the predicted probabilities $p_i$ of all prompt templates and then train on these labels with corresponding algorithms. We found that annotating partial labels achieves better results, especially under extreme circumstances when most prompt templates fail to provide satisfactory predictions. And at this time, as long as one prompt template makes a correct prediction, the prediction will be included in the candidate label set and the difficult of the algorithm to recognize it as the correct label with the help of consistency regularization is greatly decreased. This is very helpful when the characteristic of downstream task is unknown and prompt engineering can hardly be performed.

For ``image-text-to-text" models such as LLaVA, we design several prompt templates and concatenate them with all class names of the target task as choices. By incorporating the input image, we then query the model in a conversational format to elicit classification results. Specifically, the prompts are structured to guide the model to select from the provided class names or generate relevant categories, leveraging the model's multi-modal reasoning capabilities. Similarly, the pre-trained model annotates a classification category for the image based on each prompt template, and the candidate label set is obtained by taking the union of all classification results from all prompt templates.

\section{Methodology}

\subsection{Warm-up training}
Our approach first performs warm-up training for several epochs using noisy partial labels annotated by the pre-trained annotator. After the model integrates knowledge from the pre-trained "teacher", our approach fully relies on self-training based on collaborative pseudo-labeling and feature representation optimization to achieve performance improvements on the trained downstream tasks beyond those of the pre-trained model.

We adopt the partial cross-entropy loss as the supervised learning loss, enforcing the model to predict probabilities on candidate labels. Meanwhile, considering that the pre-trained model may fail to include the valid labels in the candidate label set, we hope the model can retain the possibility that non-candidate labels are the valid labels of the samples, so as to predict them as pseudo-labels in subsequent self-training stages. To this end, we adopt the negative entropy loss to prevent from over-remembering the noisy supervision. The training loss for warm-up epochs can be written as:

\begin{equation}
    L_{warm} = - \log{\sum\nolimits_{j=1}^{C} y_j f_j(x)} - \mathcal{E}(x),
\end{equation}
where $\mathcal{E}(\cdot)$ denotes the entropy function.

\subsection{Co-pseudo-labeling}
\begin{figure}
    \centering
    \includegraphics[width=0.8\linewidth]{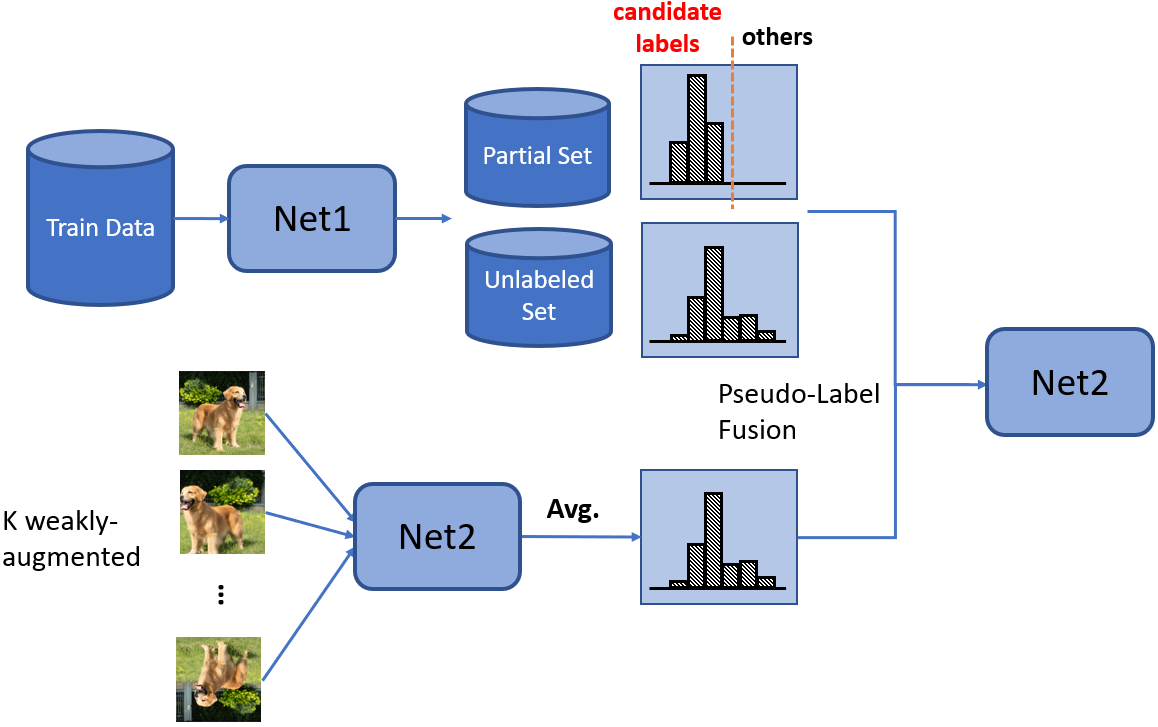}
    \caption{Schematic diagram of the Co-Pseudo-Labeling step in our method (taking the example of using the knowledge of Net1 to assist the training of Net2). We use Net1 to divide the training set into a "Partial Set" and an "Unlabeled Set" based on the credibility of the partial labels annotated by the pre-trained model. For samples in the Partial Set where the partial labels are considered trustworthy, we only retain their prediction probabilities on the candidate labels. Then, the prediction probabilities of Net1 and the prediction probabilities of Net2 itself are fused and provided to Net2 for training in the next epoch.}
    \label{co_pl}
\end{figure}

\begin{figure*}
    \centering
    \includegraphics[width=1.0\linewidth]{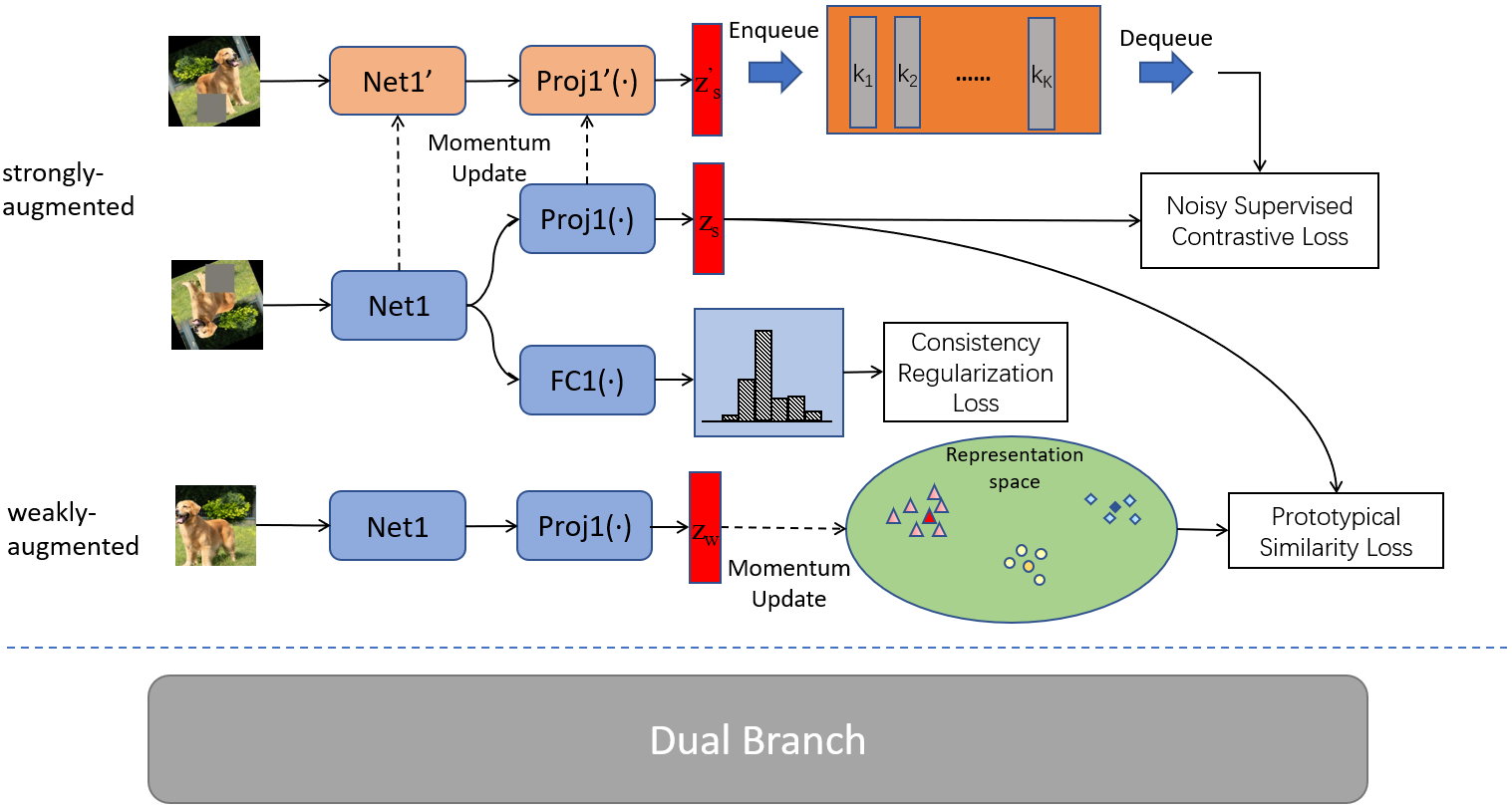}
    \caption{Schematic diagram of the self-training and feature representation optimization in our method. We use the pseudo-labels assigned by the Co-Pseudo-Labeling to perform consistency regularized training on strongly-augmented samples. Meanwhile, we use weakly-augmented samples to maintain a prototype vector for the projected feature representation of each category (shown in bold color) in a shared representation space between both networks, and enforce that the similarity distribution between the projected representations of strongly-augmented samples and the prototype vectors aligns with the predicted class distribution of these samples. Additionally, we maintain a momentum-updated network for each neural network to iteratively optimize the model's representation ability and pseudo-labels via noisy supervised contrastive learning.}
    \label{train}
\end{figure*}

Our method simultaneously trains two neural networks, denoted as Net1 $f(x; \theta_1)$ and Net2 $f(x; \theta_2)$, which collaboratively purify training labels for each other and obtain the pseudo-labels. The advantage of this approach is that it can effectively reduce the confirmation bias that can arise from mimicking the pre-trained model's behavior comparing to using self-generated pseudo-labels. In the following, we take the example of using knowledge from Net1 to provide pseudo-labels for Net2 (See Fig. \ref{co_pl}).

Firstly, we attempt to identify the provided partial labels as valid or noisy, i.e., whether the true labels are in the candidate label sets. Drawing inspiration from the minimal-loss criterion \cite{arazo2019unsupervised,chen2019understanding} which assumes that noise-free samples are easier to learn. We speculate that if the partial label of the current sample is valid, the model warm-up trained using supervised loss can predict the sample to categories within its candidate label set with a higher probability. During training, our method utilizes two types of data augmentation \cite{DBLP:conf/nips/SohnBCZZRCKL20,berthelot2019remixmatch}, i.e., weak data augmentation $\textrm{Aug}_w(\cdot)$ and strong data augmentation $\textrm{Aug}_s(\cdot)$, and aims to perform consistency regularization via instructing the training on strongly-augmented samples with guidance from their weakly-augmented variants.

Specifically, we calculate the following division loss over the predicted probabilities of all samples with weak data augmentation of Net1 $\{L_{div}(\textrm{Aug}_w(x^i); \theta_1)\}_{i=1}^{N}$.
\begin{equation}
    L_{div}(x; \theta) = - \log{f_j(x; \theta)}, j = \underset{j \in \mathcal{Y}, y_j = 1}{\arg\max} \ f_j(x; \theta),
\end{equation}
where $f_j(x; \theta)$ indicates the predicted probability on the $j$-th category of neural network with parameter $\theta$, $\mathcal{Y}$ represents the label space and $N$ is total number of training samples. Then, we use a two-component Gaussian mixture model (GMM)\cite{permuter2006study} to fit the above losses to classify the whole training set into a partial set whose partial labels annotated by the pre-train model are assumed to be valid with a probability $w$, and an unlabeled set whose annotated partial labels are assumed to be non-valid and discarded. We use $\mathcal{P} = \{(x, y, p, w) | L_{div}(x; \theta_1) < \tau_{div}\}$ to denote the partial set, where $p = (p_1, p_2, \dots, p_C)$ is the predicted label distribution of $x$ after re-scaling with Eq.\ref{rescale}, which eliminates the probabilities outside the candidate label set due to the validity of partial labels of $x^i$.
\begin{equation}
    p_j = \frac{y_j f_j(\textrm{Aug}_w(x); \theta_1)}{\sum\nolimits_{k=1}^{C} y_k f_k(\textrm{Aug}_w(x); \theta_1)}, \quad \text{for } j = 1, 2, \dots, C.
    \label{rescale}
\end{equation}

We use $\mathcal{U} = \{(x, p) | L_{div}(x; \theta_1) >= \tau_{div}\}$ to denote the unlabeled set. For samples where it is uncertain whether their partial labels are reliable, we discard the partial labels and predict the class distribution in a more cautious manner, i.e., using the average of the class distributions of $K$ weakly-augmented versions. The $p^i$ of unlabeled set can be calculated as: 
\begin{equation}
    p = \frac{1}{K} \sum_{k=1}^K f(\textrm{Aug}_w(x); \theta_{1})
\end{equation}

For more robust and generalizable pseudo-labels, we adopt pseudo-label fusion from both networks. We combine the predicted label distributions from Net1 with the average predicted probabilities of $K$ weakly-augmented inputs from Net2. The confidences of the validity of the partial labels for the samples in the partial set, i.e. $w$, are taken as the fusion weights of predicted label distributions from Net1; and the fusion weights of the other network are set as $1-w$. For samples in the unlabeled set, we set weights of the predicted probabilities of both networks to $w=0.5$. The fused probabilities can be calculated by: 
\begin{equation}
p' = w p + (1 - w) \bar{p}
\label{label_fuse}
\end{equation}
Here, $\bar{p}$ represents the average predicted probabilities of Net2. Ablation experiments show that the exploitation of the unlabeled split is of crucial importance for achieving performance improvements.

Finally, the pseudo-labels are sharpened with a temperature of $T$ to obtain more discriminative label distributions,
\begin{equation}
    \tilde{p}_j = \frac{(p'_j)^{1/T}}{\sum\nolimits_{k=1}^{C} (p'_k)^{1/T}}, \quad \text{for } j = 1, 2, \dots, C.
    \label{sharpen}
\end{equation}

\subsection{Self-training}
We use the pseudo-label distributions obtained through co-pseudo-labeling to guide the strongly-augmented samples' outputs to perform self-training. For samples in the training batch of the partial set where the model is relatively confident, we use the cross-entropy loss; while for samples in the unlabeled set, we use mean square loss due to its noise-tolerant property. The training objective can be written as:
\begin{equation}
L_{cr} = \frac{1}{|\mathcal{B}_P|} \sum_{x \in \mathcal{B}_P} \mathcal{H}(\tilde{p}; f(\textrm{Aug}_s(x))) + \frac{\lambda_u}{C \cdot |\mathcal{B}_U|} \sum_{x \in \mathcal{B}_U} ||\tilde{p} - f(\textrm{Aug}_s(x))||^2_2,
\end{equation}
where $\mathcal{B}_P$ and $\mathcal{B}_U$ stand for the training batch of samples of the partial set $\mathcal{P}$ and the unlabeled set $\mathcal{U}$, respectively; $\mathcal{H}(\cdot; \cdot)$ is the cross-entropy loss, $||\cdot||^2_2$ denotes the squared L2-norm and $\lambda_u$ is the weight parameter.

\subsection{Prototypical similarity alignment}
To further enhance the model's representational capacity for specific tasks, while reinforcing the predictive consistency between the label space and the representation space, we employ the prototypical similarity alignment, in which the similarity distribution measured by representative class representations is aligned to the predictive label distribution.

During the training process, we maintain a cluster center for each category in a projected representation space shared by two networks (See Fig.\ref{train}), called ``prototype", denoted by $\{o^*_j\}_{j=1}^C$. The "prototype" could be considered as the representative representation of $N_z$ data points $\{z^i\}_{i=1}^{N_z}$ that belong to the same class in the shared space, which has the minimal expected distance with them.
\begin{equation}
    o^* = \arg\min_{o} \mathbb{E}_{z \sim \mathcal{Z}}[d(z, o)],
\label{prototype_1}
\end{equation}
in which $\mathcal{Z}$ represents the discrete probability distribution over $N_z$ data points $\{z^i\}_{i=1}^{N_z}$, and $d(\cdot, \cdot)$ is the distance function which measures the divergence between two vectors. Since that the ground-truth labels are unknown, we extends Eq.\ref{prototype_1} and derives prototype $o^*_j$ for class $j$ by via minimizing the expected distance between the class prototype and embeddings of the instances associated with that class under the conditional probability $z \sim p(z|j)$, obtaining:
\begin{equation}
    o^*_j = \arg\min_{o} \mathbb{E}_{z \sim p(z|j)}[d(z, o)].
\label{prototype_2}
\end{equation}

Following previous works \cite{aaai2026collaborative}, we use the squared Euclidean distance as the instantiation of function $d(\cdot, \cdot)$ due to its excellent properties as a regular Bregman divergence \cite{banerjee2005optimality}. In detail, the regular Bregman divergences denote a particular class of distance functions satisfying the following equation:
\begin{equation}
    d_{\phi}(a,b) = \phi(a) - \phi(b) - \langle a-b, \nabla_{\phi}(b) \rangle,
\end{equation}
where $a,b \in \mathbb{R}^{d'}$, and $\phi(\cdot)$ is a differentiable and strictly convex function w.r.t. $d_{\phi}(\cdot,\cdot)$, while $\langle \cdot,\cdot \rangle$ refers to the inner product function, and $\nabla_{\phi}(\cdot)$ denotes the gradient function of $\phi(\cdot)$. Then, we can derive the following proposition:

\begin{proposition}
Given the bregman divergence $d(a, b) = d_{\phi}(a, b) = ||a - b||^2$, the prototype of each class $j$ has a unique formulation to minimize the problem of Eq.\ref{prototype_2}, which is given by $o^*_j = \mathbb{E}_{z \sim p(z|j)}[z]$
\end{proposition}

\begin{proof}
For each class $j$, let $J_{\phi}(o) = \mathbb{E}_{z \sim p(z|j)}[d_{\phi}(z,o)] = \sum_{z_i \in \mathcal{B}} p(z_i|j) d_{\phi}(z_i,o)$ denote the function we are trying to minimize, and $o^*_j = \mathbb{E}_{z \sim p(z|j)}[z] = \sum_{z_i \in \mathcal{B}} p(z_i|j) z_i$. Then for $\forall o \in \mathcal{R}^{d'}$, we have:
\begin{align}
    \notag
    J_{\phi}(o) - J_{\phi}(o^*_j) &= \sum_{z_i \in \mathcal{B}} p(z_i|j) d_{\phi}(z_i,o) \\ \notag
    &\quad- \sum_{z_i \in \mathcal{B}} p(z_i|j) d_{\phi}(z_i,o^*_j) \\ \notag
    &= \phi(o^*_j) - \phi(o) \\ \notag
    &\quad- \langle \sum_{z_i \in \mathcal{B}} p(z_i|j) z_i - o, \nabla_{\phi}(o)\rangle \\ \notag
    &\quad+ \langle \sum_{z_i \in \mathcal{B}} p(z_i|j) z_i - o^*_j, \nabla_{\phi}(o^*_j)\rangle \\ \notag
    &= \phi(o^*_j) - \phi(o) - \langle o^*_j - o, \nabla_{\phi}(o) \rangle \\
    &= d_{\phi}(o^*_j, o) \geq 0.
\end{align}
Given $\phi(\cdot)$ is strictly convex, thus $o^*_j$ is the unique minimizer of $J_{\phi}$.
\end{proof}

Based on the above proposition, the calculation formula for each prototype $o^*_j$ is derived as follows:
\begin{equation}
    o^*_j = \mathbb{E}_{z \sim p(z|j)} [z] = \sum_{z_i \in \mathcal{B}} p(z_i|j) z_i.
\end{equation}
According to the Bayes’ theorem, we have:
\begin{equation}
    o^*_j = \sum_{z_i \in \mathcal{B}} p(j|z_i) \frac{p(z_i)}{p(j)} z_i.
\end{equation}

Based on the assumption that the instances are uniformly distributed, i.e. for every sample $z_i \in \mathcal{B}$, $p(z_i) = \frac{1}{|\mathcal{B}|}$, we can calculate each prototype $o^*_j$ by using the estimated label distribution of co-pseudo-labeling $\tilde{p}^i_j$ as the conditional probability $p(j|z_i)$:
\begin{align}
    \notag
    o^*_j &= \frac{p(z_i)}{p(j)} \sum_{z_i \in \mathcal{B}} p(j|z_i) z_i, \\ \notag
    &= \frac{1}{|\mathcal{B}| \cdot p(j)} \sum_{z_i \in \mathcal{B}} \tilde{p}^i_j z_i, \\
    &= \frac{1}{q} \sum_{z_i \in \mathcal{B}} \tilde{p}^i_j z_i.
\end{align}
where $q = \sum_{z_i \in \mathcal{B}} \tilde{p}^i_j$ stands for the sum of predicted probabilities of class $j$ within the current batch.

For stability, the class prototypes are momentum updated during training with the projected features of weakly-augmented samples of partial batches with Eq.\ref{update}.
\begin{equation}
    o^*_j \leftarrow \gamma o^*_j + (1 - \gamma) \frac{1}{q} \sum_{z_i \in \mathcal{B}_p} \tilde{p}^i_j z_i,
    \label{update}
\end{equation}

For current image $x^i$, we obtain the output representations of the strongly-augmented image $\textrm{Aug}_s(x^i)$ of the two neural networks, respectively, which are then projected to the shared representation space $z^i_s = g(f(\textrm{Aug}_s(x^i); \bar{\theta}))$, in which $g(\cdot)$ represents the MLP projector and $\bar{\theta}$ represents the neural network parameters $\theta$, excluding the last fully-connected layer. In implementation, the projectors of both networks are implemented by a two-layer MLP with L2 normalization.

Then, our method calculate the similarity distribution over the projected representation of current image $z^i_s$ and class prototypes $\{o^*_j\}_{j=1}^C$ as $s^i = \textrm{softmax}(z^i_s o^*_1, z^i_s o^*_2, \dots, z^i_s o^*_C)$, which is then aligned to its pseudo-label distribution $\tilde{p^i}$ to enforce consistency. Similarly, we choose KL-Divergence with temperature $T'$ for samples in the partial set, whose pseudo-labels have much higher accuracies and mean square error for samples in the unlabeled set. The loss functions for prototypical similarity alignment can be written as:
\begin{equation}
    L_{prot} = \frac{1}{|\mathcal{B}_P|} \sum_{x \in \mathcal{B}_P} D_{\text{KL}}(\tilde{p}/T' || s/T') + \frac{\lambda_u}{C \cdot |\mathcal{B}_U|} \sum_{x \in \mathcal{B}_U} ||\tilde{p} - s||^2_2,
\end{equation}
where $D_{\text{KL}}(\cdot || \cdot)$ stands for the KL-Divergence.

\subsection{Noisy contrastive learning}
To further exploit from the data distribution property of downstream unlabeled images while enhancing the model's representation ability, we employ the widly-used contrastive learning while alleviating the negative impact of noisy supervision to the great extent. 

We utilize contrastive learning to pull together representations of samples from the same class while pushing apart those from different classes, enabling the model to encode more discriminative features on downstream data. In implementation, we adopt the MoCo \cite{DBLP:conf/cvpr/He0WXG20} framework, in which a large-size "first-in-first-out" queue of representations of strong-augmented images encoded by the momentum updated copy of our model is maintained. We select positive and negative set regarding the current image representation from the representation queue according to their pseudo-labels, and optimize the following noisy-tolerant contrastive loss:
\begin{equation}
\begin{aligned}
\mathcal{L}_{ncont} =& - \frac{1}{|\mathcal{B}_P|+|\mathcal{B}_U|} \sum_{x \in \mathcal{B}_P \cup \mathcal{B}_U} \frac{1}{|P(x)|} \sum\nolimits_{z_+ \in P(x)} \\ 
&\log \frac{\exp(z^\top_s z_+ / T'')}{\sum\limits_{z_+ \in P(x)} \exp(z^\top_s z_+ / T'') + \sum\limits_{z_- \in N(x)} \exp(z^\top_s z_- / T'')},
\end{aligned}
\end{equation}
where $P(x)$ and $N(x)$ separately denote the set of selected positive and negative examples for image $x$, $T'' \geq 0$ is the temperature. We treat $x \in \mathcal{P}$ as confident samples and $x \in \mathcal{U}$ as lacking of confidence and applying the selection strategy for $P(x)$ and $N(x)$ in \cite{wang2024controller}.

Finally, the overall training loss is:
\begin{equation}
    L = L_{cr} + \beta_1 L_{prot} + \beta_2 \mathcal{L}_{ncont},
\end{equation}
where $\beta_1$ and $\beta_2$ are weight parameters.

\subsection{Implementation details}
\subsubsection{Data augmentation}
In weak data augmentation, we randomly shift the original image by up to 12.5\% in all directions, followed by a random horizontal flip. In strong data augmentation, we first perform random cropping and random horizontal flipping on the image (as in weak data augmentation), followed by RandAugment \cite{cubuk2020randaugment}. In RandAugment, we first randomly apply one of the image processing functions preset by Python image library (PIL), such as AutoContrast, Rotate and Sharpness, and then execute cutout.

\subsubsection{MixUp}
For further facilitating the effectiveness and robustness of consistency regularized training, we adopt the MixUp \cite{zhang2017mixup,li2020dividemix} technique, where each sample is interpolated with another sample randomly chosen from the combined mini-batch of the partial set and the unlabeled set with the following equations. 
\begin{equation}
    \lambda \sim Beta(\alpha, \alpha),
\end{equation}
\begin{equation}\label{mixup}
    \lambda' = \max (\lambda, 1-\lambda),
\end{equation}
\begin{equation}
    x_{mix} = \lambda' x^1 + (1 - \lambda') x^2,
\end{equation}
\begin{equation}
    \tilde{p}_{mix} = \lambda' \tilde{p}^1 + (1 - \lambda') \tilde{p}^2.
\end{equation}

\subsubsection{Positive and negative set selection}
In noisy contrastive learning, we select positive and negative sample sets for each current image representation to compute the contrastive learning loss. As the contrastive loss is optimized, the representation of the current image is pulled closer to those in the positive set while being pushed farther from those in the negative set. To ensure the accuracy of the positive set while maximizing the size of the negative set, which has been demonstrated as crucial in previous studies \cite{DBLP:conf/cvpr/He0WXG20,chen2020simple}, our selection strategy is as follows:

\begin{itemize}
    \item For a confident input, confident samples with the same class prediction are chosen as positives, while all other samples with different predictions are treated as negatives;

    \item For an unconfident input, only the representation of another augmented variant of the input is selected as the positive. Consistent with the case of confident inputs, all samples with different predictions are considered negatives.
\end{itemize}

\section{Experiments}
\subsection{Experimental setup}
\subsubsection{Experimental datasets and VLMs}
We conduct experiments on six image classification benchmark datasets, including general-purpose image datasets CIFAR-10, CIFAR-100 \cite{krizhevsky2009learning}, SVHN \cite{goodfellow2013multi}, Fashion-MNIST \cite{xiao2017fashion}, as well as fine-grained datasets EuroSAT \cite{helber2019eurosat} and GTSRB \cite{houben2013detection}.

We annotate the images of these datasets with prevailing pre-trained VLMs including: CLIP ViT-B/32, CLIP ViT-B/16 \cite{radford2021learning} and LLaVA-1.5 \cite{liu2023visual}. The class names for prompting VLMs are manually assigned and are the same for all comparison methods. The detailed information of these datasets and pre-trained VLMs can be found in \ref{appendix-datasets}.

\begin{table*}[]
\centering
\setlength{\tabcolsep}{3pt}
\small
\begin{tabular}{@{}l|*{6}{>{\centering\arraybackslash}p{0.11\textwidth}}@{}}
\toprule
Methods & CIFAR-10 & CIFAR-100 & SVHN & F-MNIST & EuroSAT & GTSRB \\ \midrule
Partial Acc. & 95.21\% & 78.50\% & 38.39\% & 77.65\% & 67.11\% & 41.78\% \\
Avg. num     & 1.39 & 2.36 & 2.41 & 1.58 & 3.26 & 2.84 \\ \midrule
Zero-Shot (train)  & 88.40\% & 61.83\% & 9.34\% & 62.57\% & 32.26\% & 24.87\% \\
Zero-Shot* (train) & 89.09\% & 62.75\% & 9.33\% & 65.81\% & 30.61\% & 25.53\% \\
Zero-Shot (test)   & 88.51\% & 61.55\% & 8.63\% & 61.46\% & 31.49\% & 25.14\% \\
Zero-Shot* (test)  & 89.01\% & 62.74\% & 8.82\% & 65.14\% & 30.78\% & 25.57\% \\
KD$_{unsup}$       & 87.39\% & 56.31\% & 8.01\% & 65.91\% & 35.24\% & 25.70\% \\
KD$_{unsup}$*      & 87.74\% & 56.80\% & 8.21\% & 67.60\% & 34.13\% & 26.98\% \\
DivideMix          & 93.32\% & 65.76\% & 16.46\% & 71.60\% & 42.41\% & 30.07\% \\
DivideMix*         & 93.83\% & 66.03\% & 17.16\% & 74.21\% & 37.74\% & 32.69\% \\
CR-DPLL            & 84.20\% & 60.05\% & 6.82\% & 71.27\% & 8.85\% & 27.16\% \\
ALIM-Onehot        & 93.18\% & 64.60\% & 17.06\% & 72.42\% & 34.57\% & 31.06\% \\
ALIM-Scale         & 93.59\% & 64.61\% & 20.66\% & 72.36\% & 37.11\% & 31.81\% \\
Ours               & \textbf{94.06}\% & \textbf{71.04}\% & \textbf{46.57}\% & \textbf{76.28}\% & \textbf{65.54}\% & \textbf{41.18}\% \\ \midrule
CoOp 1-shot & 74.25\% & 46.08\% & 15.10\% & 70.42\% & 53.74\% & 22.53\% \\
CoOp 2-shot & 75.19\% & 46.14\% & 20.90\% & 72.77\% & 60.44\% & 20.01\% \\
CoOp 4-shot & 75.66\% & 48.41\% & 28.18\% & 76.10\% & 68.52\% & 21.02\% \\
CoOp 8-shot & 75.00\% & 52.18\% & 27.14\% & 78.99\% & 75.78\% & 25.55\% \\
CoOp 16-shot & 74.90\% & 52.18\% & 26.12\% & 77.86\% & 76.09\% & 22.09\% \\
Ours 1-shot & 92.87\% & 69.96\% & 29.66\% & 70.11\% & 71.21\% & 53.67\% \\
Ours 2-shot & 93.20\% & 70.05\% & 42.26\% & 73.24\% & 80.80\% & 60.12\% \\
Ours 4-shot & 93.15\% & 70.28\% & 56.30\% & 80.42\% & 85.50\% & 66.12\% \\
Ours 8-shot & 93.42\% & 70.12\% & 65.08\% & 84.93\% & 89.04\% & 75.83\% \\
Ours 16-shot & 94.10\% & 71.12\% & 80.67\% & 87.88\% & 91.98\% & 87.88\% \\
\bottomrule
\end{tabular}
\caption{Accuracy comparisons on CLIP ViT-B/32 annotated datasets. Best performances are shown in bold.}
\label{main_clip_b32}
\end{table*}

\begin{table*}[]
\centering
\setlength{\tabcolsep}{3pt}
\small
\begin{tabular}{@{}l|*{6}{>{\centering\arraybackslash}p{0.11\textwidth}}@{}}
\toprule
Methods & CIFAR-10 & CIFAR-100 & SVHN & F-MNIST & EuroSAT & GTSRB \\ \midrule
Partial Acc. & 95.56\% & 82.15\% & 58.84\% & 88.03\% & 65.96\% & 41.45\% \\
Avg. num     & 1.31 & 2.05 & 2.23 & 2.00 & 2.91 & 2.67 \\ \midrule
Zero-Shot (train)  & 89.63\% & 65.28\% & 36.19\% & 66.82\% & 35.72\% & 32.41\% \\
Zero-Shot* (train) & 90.25\% & 66.04\% & 33.52\% & 68.87\% & 36.52\% & 32.29\% \\
Zero-Shot (test)   & 89.22\% & 64.54\% & 40.09\% & 66.59\% & 35.34\% & 32.57\% \\
Zero-Shot* (test)  & 89.85\% & 65.46\% & 36.60\% & 68.84\% & 36.77\% & 32.38\% \\
KD$_{unsup}$       & 87.34\% & 56.71\% & 41.26\% & 70.48\% & 37.72\% & 33.12\% \\
KD$_{unsup}$*      & 87.50\% & 58.43\% & 37.12\% & 71.54\% & 40.76\% & 32.60\% \\
DivideMix          & 85.49\% & 69.18\% & 46.07\% & 71.17\% & 53.76\% & 44.23\% \\
DivideMix*         & 86.49\% & 69.17\% & 37.23\% & 75.36\% & 49.02\% & 42.53\% \\
CR-DPLL            & 93.63\% & 61.56\% & 34.52\% & 74.59\% & 38.48\% & 33.24\% \\
ALIM-Onehot        & 93.79\% & 65.94\% & 45.72\% & 75.92\% & 49.72\% & 33.71\% \\
ALIM-Scale         & 94.25\% & 67.39\% & 47.50\% & 73.87\% & 48.94\% & 34.74\% \\
Ours               & \textbf{94.38}\% & \textbf{72.03}\% & \textbf{67.00}\% & \textbf{77.27}\% & \textbf{64.28}\% & \textbf{50.74}\% \\ \midrule
CoOp 1-shot & 77.24\% & 44.06\% & 20.08\% & 62.33\% & 47.30\% & 28.39\% \\
CoOp 2-shot & 79.05\% & 47.82\% & 40.45\% & 67.53\% & 56.74\% & 28.24\% \\
CoOp 4-shot & 78.36\% & 50.39\% & 39.18\% & 70.26\% & 66.85\% & 25.43\% \\
CoOp 8-shot & 78.95\% & 51.83\% & 44.69\% & 73.36\% & 73.37\% & 19.74\% \\
CoOp 16-shot & 80.14\% & 55.50\% & 45.56\% & 74.89\% & 76.48\% & 26.61\% \\
Ours 1-shot & 93.06\% & 70.02\% & 65.26\% & 76.97\% & 72.85\% & 60.89\% \\
Ours 2-shot & 93.50\% & 70.81\% & 70.70\% & 79.22\% & 80.06\% & 63.13\% \\
Ours 4-shot & 94.02\% & 71.34\% & 86.36\% & 82.45\% & 86.33\% & 75.90\% \\
Ours 8-shot & 94.20\% & 70.96\% & 91.53\% & 85.49\% & 89.74\% & 86.44\% \\
Ours 16-shot & 94.45\% & 71.90\% & 91.53\% & 87.67\% & 93.24\% & 96.71\% \\
\bottomrule
\end{tabular}
\caption{Accuracy comparisons on CLIP ViT-B/16 annotated datasets. Best performances are shown in bold.}
\label{main_clip_b16}
\end{table*}

\begin{table*}[]
\centering
\setlength{\tabcolsep}{3pt}
\small
\begin{tabular}{@{}l|*{6}{>{\centering\arraybackslash}p{0.11\textwidth}}@{}}
\toprule
Methods & CIFAR-10 & CIFAR-100 & SVHN & F-MNIST & EuroSAT & GTSRB \\ \midrule
Partial Acc. & 94.83\% & 60.23\% & 82.74\% & 54.03\% & 63.26\% & 46.45\% \\
Avg. num     & 1.12 & 1.34 & 2.40 & 1.80 & 1.85 & 2.67 \\ \midrule
Zero-Shot (train)  & 89.64\% & 49.43\% & 51.94\% & 46.93\% & 44.77\% & 37.36\% \\
Zero-Shot* (train) & 91.80\% & 51.06\% & 53.76\% & 42.53\% & 45.68\% & 34.03\% \\
Zero-Shot (test)   & 89.56\% & 48.90\% & 52.35\% & 47.50\% & 44.78\% & 37.40\% \\
Zero-Shot* (test)  & 91.56\% & 50.87\% & 54.20\% & 42.64\% & 45.70\% & 34.05\% \\
KD$_{unsup}$       & 83.72\% & 44.13\% & 64.21\% & 47.90\% & 51.67\% & 32.91\% \\
KD$_{unsup}$*      & 87.24\% & 41.23\% & 68.23\% & 45.78\% & 49.48\% & 34.96\% \\
DivideMix          & 83.59\% & 55.52\% & 73.33\% & 52.33\% & 62.56\% & 46.37\% \\
DivideMix*         & 84.60\% & 56.81\% & 73.90\% & 51.26\% & 62.95\% & 45.85\% \\
CR-DPLL            & 91.27\% & 33.50\% & 42.65\% & 32.22\% & 48.13\% & 24.28\% \\
ALIM-Onehot        & 92.63\% & 41.28\% & 66.65\% & 46.75\% & 51.26\% & 46.75\% \\
ALIM-Scale         & 93.08\% & 43.83\% & 66.08\% & 47.30\% & 51.43\% & 35.58\% \\
Ours               & \textbf{94.47}\% & \textbf{62.54}\% & \textbf{86.70}\% & \textbf{66.28}\% & \textbf{77.24}\% & \textbf{47.90}\% \\ \midrule
LoRA 1-shot & 87.59\% & 60.46\% & 55.98\% & 58.26\% & 60.15\% & 38.87\% \\
LoRA 2-shot & 86.80\% & 65.07\% & 65.85\% & 62.76\% & 70.93\% & 44.78\% \\
LoRA 4-shot & 89.54\% & 70.13\% & 74.04\% & 70.43\% & 72.67\% & 40.70\% \\
LoRA 8-shot & 91.25\% & 73.91\% & 77.61\% & 75.38\% & 83.44\% & 49.69\% \\
LoRA 16-shot & 93.84\% & 75.00\% & 77.88\% & 76.44\% & 84.35\% & 50.70\% \\
Ours 1-shot & 93.21\% & 58.04\% & 88.60\% & 73.16\% & 79.57\% & 57.75\% \\
Ours 2-shot & 93.97\% & 60.97\% & 94.01\% & 78.29\% & 82.92\% & 70.04\% \\
Ours 4-shot & 94.10\% & 62.39\% & 96.22\% & 81.78\% & 81.26\% & 89.19\% \\
Ours 8-shot & 94.50\% & 64.75\% & 95.80\% & 83.74\% & 90.93\% & 95.88\% \\
Ours 16-shot & 94.72\% & 66.58\% & 96.44\% & 85.73\% & 93.26\% & 98.27\% \\
\bottomrule
\end{tabular}
\caption{Accuracy comparisons on LLaVA-1.5 annotated datasets. Best performances are shown in bold.}
\label{main_llava}
\end{table*}

\subsubsection{Comparing methods}
We compare the performances of our method with various types of NPLL methods under partial label annotations: CR-DPLL \cite{wu2022revisiting}, ALIM-Onehot and ALIM-Scale \cite{xu2024alim}, in which CR-DPLL aims to learn from clean partial labels and ALIM-Onehot and ALIM-Scale are able to deal with noisy candidates. We also compare with DivideMix \cite{li2020dividemix}, which is a state-of-the-art noisy label learning method, using single labels annotated by VLMs. 

We also compare our approach with three widely-adopted pre-trained model application paradigms: 
\begin{itemize}
    \item \textbf{Zero-shot inference} operates by leveraging the pre-trained model’s intrinsic knowledge and natural language task descriptions (e.g., instructional prompts or example demonstrations) to directly infer outputs for unseen tasks, eliminating the need for task-specific training data through semantic alignment between model parameters and task semantics; 

    \item \textbf{Unsupervised knowledge distillation} (KD) transfers knowledge from a large "teacher" model to a smaller "student" model in an unlabeled data setting, where the student model is optimized to mimic the teacher’s output distributions (e.g., class logits or hidden-layer representations) without access to explicit task-relevant labels, aiming to improve efficiency or adapt to resource-constrained environments;

    \item \textbf{Few-shot fine-tuning} refers to fine-tuning models with minimal labeled examples while typically freezing most pre-trained parameters and introducing minimal trainable parameters, exemplified by prompt learning, which constructs task-specific textual templates and adapts prompt parameters or output layers to transform tasks into language-model-friendly formats, and LoRA (Low-Rank Adaptation), which fine-tunes only added low-rank matrix parameters to enable efficient adaptation with reduced computational costs.
\end{itemize}

For zero-shot inference, unsupervised KD and DivideMix, which learn from single labels, we record the performances with single prompt template as well as using the average of predicted probabilities of multiple prompt templates (superscript with asterisks for distinction) for comprehensive comparison.

Zero-shot inference, unsupervised KD and weakly-supervised methods require no human labeling while few-shot fine-tuning is performed with 1, 2, 4, 8, and 16 labeled samples per class. We choose CoOp \cite{zhou2022learning} as the prompt learning comparing method for CLIP ViT-B/32 and CLIP ViT-B/16, and use LoRA \cite{hu2022lora} for LLaVA-1.5. Meanwhile, we extend our approach to the semi-supervised learning scenario, where the model is trained using a small number of manually annotated true label samples and noisy partial labels annotated by CLIP or LLaVA, and compare the performances with the few-shot fine-tuning methods.

The average amount of candidate labels per training sample (denote as Avg. num) and the proportions of ground-truth label being inside of the candidate sets (denote as Partial Acc.) is also recorded for partially annotated datasets.

\subsubsection{Implementation details}
We use the PreAct ResNet-18 \cite{he2016identity} as the backbone for all comparing methods. The training batch-size is 256, and the number of warm up and total epochs are chosen from 20 to 100 and 100 to 800, respectively. The threshold for dividing partial and unlabeled set is set as $\tau_{div}=0.5$. The number of weakly-augmented inputs for co-pseudo-labeling is $K = 2$ and the sharpening temperature of pseudo-labels is $T = 0.5$. The weight parameter $\lambda_u$ is set to 1. The dimension of projected representations is $d'=128$, and the length of feature representation queue updated by momentum encoder is 8192. The experiments are all carried on NVIDIA V100 / 3090 GPUs.

\begin{table*}[t]
\centering
\setlength{\tabcolsep}{6pt}
\resizebox{1.0\textwidth}{!}{
\begin{tabular}{@{}l|ccccccccc@{}}
\toprule
\multirow{2}{*}{\centering Methods} & \multicolumn{3}{c}{$q=0.01$}  & \multicolumn{3}{c}{$q=0.03$}  & \multicolumn{3}{c}{$q=0.05$}  \\ \cmidrule(l){2-10} 
                           & $\eta=0.1$   & $\eta=0.2$   & $\eta=0.3$   & $\eta=0.1$   & $\eta=0.2$   & $\eta=0.3$   & $\eta=0.1$   & $\eta=0.2$   & $\eta=0.3$   \\ \midrule
CC                         & 53.63\% & 48.84\% & 45.50\% & 51.85\% & 47.48\% & 43.37\% & 50.64\% & 45.87\% & 40.87\% \\
RC                         & 52.73\% & 48.59\% & 45.77\% & 52.15\% & 48.25\% & 43.92\% & 46.62\% & 45.46\% & 40.31\% \\
LWC                        & 53.16\% & 48.64\% & 45.51\% & 51.69\% & 47.60\% & 43.39\% & 50.55\% & 45.85\% & 39.83\% \\
LWS                        & 56.05\% & 50.66\% & 45.71\% & 53.59\% & 48.28\% & 42.20\% & 45.46\% & 39.63\% & 33.60\% \\
PiCO                       & 68.27\% & 62.24\% & 58.97\% & 67.38\% & 62.01\% & 58.64\% & 67.52\% & 61.52\% & 58.18\% \\
CR-DPLL                     & 68.12\% & 65.32\% & 62.94\% & 67.53\% & 64.29\% & 61.79\% & 67.17\% & 64.11\% & 61.03\% \\
PiCO+                      & 75.04\% & 74.31\% & 71.79\% & 74.68\% & 73.65\% & 69.97\% & 73.06\% & 71.37\% & 67.56\% \\
IRNet                      & 71.17\% & 70.10\% & 68.77\% & 71.01\% & 70.15\% & 68.18\% & 70.73\% & 69.33\% & 68.09\% \\
ALIM-Scale                 & 77.37\% & 76.81\% & 76.45\% & 77.60\% & 76.63\% & \textbf{75.92\%} & 76.86\% & \textbf{76.44\%} & \textbf{75.67\%} \\
ALIM-Onehot                & 76.52\% & 76.55\% & 76.09\% & 77.27\% & 76.29\% & 75.29\% & \textbf{76.87\%} & 75.23\% & 74.49\% \\
Co-Reg                     & \textbf{78.13\%} & \textbf{78.01\%} & \textbf{77.20\%} & \textbf{77.16\%} & \textbf{76.85\%} & 75.71\% & 76.30\% & 74.91\% & 73.45\% \\ \bottomrule
\end{tabular}}
\caption{Accuracy comparisons on synthetic NPLL datasets, best performances in bold.}
\label{synthetic}
\end{table*}

\subsection{Main results}
In the main experimental results using pre-trained VLMs (CLIP ViT-B/32, CLIP ViT-B/16, and LLaVA-1.5) to annotate unlabeled samples with noisy partial labels, our Co-Reg method consistently outperforms state-of-the-art NPLL and knowledge distillation baselines across all datasets (Table \ref{main_clip_b32},\ref{main_clip_b16},\ref{main_llava}). For instance, on CLIP ViT-B/32-annotated CIFAR-100, our method achieves 71.04\% accuracy, significantly surpassing Zero-Shot inference 62.74\%, unsupervised KD 56.80\%, noisy label method DivideMix* 66.03\% under ensemble prompting and NPLL methods like ALIM-Scale 64.61\%. The superiority is more pronounced on datasets with lower Partial Acc. or higher label noise complexity, demonstrating Co-Reg’s effectiveness in mitigating instance-dependent noise from VLMs. On LLaVA-1.5-annotated SVHN, our method achieves 86.70\% accuracy under the condition that the Partial Acc. of the pre-trained annotated candidate sets is only 82.74\%, demonstrating our method's ability to find out the correct label outside of the candidates. On LLaVA-1.5-annotated GTSRB, our method achieves 47.90\% accuracy, performing comparably to LoRA 16-shot 50.70\% that uses 16 manually annotated true labels per class, highlighting its advantages when without human-labeled data.

Notably, traditional unsupervised KD typically yields minor performance gains for downstream tasks, as it merely mimics a teacher model’s output distributions without addressing annotation noise or structural conflicts in labels. In contrast, our method incorporates NPLL into the KD framework, leveraging collaborative consistency regularization and pseudo-label purification to correct VLM-generated annotation errors. This strategies enable effective knowledge transfer while mitigating instance-dependent noise inherent in VLMs’ label predictions, leading to substantial accuracy improvements over vanilla KD without requiring additional human-labeled data.

\subsection{Semi-supervised settings}
In semi-supervised learning scenarios, we integrate a small number of manually annotated true labels (1–16 shots per class) with VLM-generated noisy partial labels. Our method substantially outperforms few-shot fine-tuning methods, i.e. CoOp for CLIP, LoRA for LLaVA, across all shot counts, VLMs and datasets except LLaVA-1.5-annotated CIFAR-100 (Table \ref{main_clip_b32},\ref{main_clip_b16},\ref{main_llava}). For example, on LLaVA-1.5’s 1-shot setting for GTSRB, our method achieves 57.75\% accuracy, far exceeding LoRA’s 38.87\%. As shot counts increase to 16, our method reaches 98.27\% on GTSRB, outperforming LoRA’s 50.70\% by 47.57\%, and has an accuracy difference of less than 1\% with training with full supervision. On LLaVA-1.5-annotated SVHN, our 1-shot result 88.80\% surpasses LoRA 1-shot 55.98\% by 32.82\%, and our 16-shot result 96.44\% surpasses LoRA 16-shot 77.88\% by 18.56\%, demonstrating that leveraging noisy partial labels alongside minimal true labels enables stronger generalization than pure few-shot fine-tuning. The performances also validate the effectiveness of our collaborative label purification and consistency regularized training in low-label regimes, where traditional methods struggle with limited supervision.

\begin{table*}[t]
\centering
\label{tab:ablation_experiment}
\setlength{\tabcolsep}{3pt}
\renewcommand{\arraystretch}{0.9}
\small
\resizebox{1.0\textwidth}{!}{
\begin{tabular}{@{}l|l|*{6}{>{\centering\arraybackslash}m{0.11\textwidth}}@{}}
\toprule
VLMs & Methods &
CIFAR-10 & CIFAR-100 & SVHN & F-MNIST & EuroSAT & GTSRB \\
\midrule
\multirow{5}{*}{CLIP ViT-B/32}
 & w/o Co-PL  & 93.54\% & 68.40\% & 42.35\% & 72.33\% & 62.00\% & 37.79\% \\
 & w/ SupCont & 93.50\% & 67.96\% & 45.05\% & 74.18\% & 63.85\% & 39.60\% \\
 & w/o proto  & 94.01\% & 69.81\% & 45.44\% & 75.50\% & 64.69\% & 40.15\% \\
 & w/o U      & 93.86\% & 65.32\% & 41.60\% & 73.33\% & 62.98\% & 38.64\% \\
 & Co-Reg     & \textbf{94.06\%} & \textbf{71.04\%} & \textbf{46.57\%} & \textbf{76.28\%} & \textbf{65.54\%} & \textbf{41.18\%} \\
\cmidrule(lr){1-8}

\multirow{5}{*}{CLIP ViT-B/16}
 & w/o Co-PL  & 93.21\% & 69.48\% & 64.36\% & 76.51\% & 62.15\% & 47.16\% \\
 & w/ SupCont & 93.46\% & 68.95\% & 65.12\% & 75.29\% & 62.58\% & 48.00\% \\
 & w/o proto  & 93.95\% & 70.56\% & 65.36\% & 77.21\% & 63.33\% & 47.59\% \\
 & w/o U      & 94.25\% & 66.30\% & 63.86\% & 76.30\% & 60.31\% & 45.90\% \\
 & Co-Reg     & \textbf{94.38\%} & \textbf{72.03\%} & \textbf{67.00\%} & \textbf{77.27\%} & \textbf{64.28\%} & \textbf{50.74\%} \\
\cmidrule(lr){1-8}

\multirow{5}{*}{LLaVA-1.5}
 & w/o Co-PL  & 93.69\% & 59.75\% & 83.47\% & 64.12\% & 74.86\% & 45.80\% \\
 & w/ SupCont & 93.00\% & 59.76\% & 83.88\% & 65.43\% & 75.08\% & 45.87\% \\
 & w/o proto  & 94.11\% & 61.50\% & 84.96\% & 65.55\% & 75.62\% & 46.31\% \\
 & w/o U      & 94.30\% & 58.96\% & 84.31\% & 64.96\% & 73.65\% & 42.18\% \\
 & Co-Reg     & \textbf{94.47\%} & \textbf{62.54\%} & \textbf{86.70\%} & \textbf{66.28\%} & \textbf{77.24\%} & \textbf{47.90\%} \\
\bottomrule
\end{tabular}}
\caption{Ablation experiments on different degenerations of our method on multiple pre-trained VLM annotated datasets. Co-Reg denotes the complete method (bold font indicates the baseline performance). All results are reported as accuracy (\%).}
\end{table*}

\subsection{Synthetic datasets}

We also conduct the experiments on synthetic datasets of CIFAR-100, following the generation process used by the previous method \cite{xu2024alim}. First, we generate partially labeled datasets by flipping negative labels $\bar{y} \neq y$ to false positive labels with a probability $q = P(\bar{y} \in Y | \bar{y} \neq y)$. Then, we generate noisy partially labeled datasets by randomly substituting the ground-truth label with a non-candidate label with a probability $\eta = P(y \notin Y)$ for each sample. We choose the noise level $\eta$ from \{0.1, 0.2, 0.3\}, and consider $q \in \{0.01, 0.03, 0.05\}$ for CIFAR-100.

We compare our method with ten PLL and NPLL methods, i.e. CC \cite{DBLP:conf/nips/FengL0X0G0S20}, RC \cite{DBLP:conf/nips/FengL0X0G0S20}, LWC \cite{DBLP:conf/icml/WenCHL0L21}, LWS \cite{DBLP:conf/icml/WenCHL0L21}, PiCO \cite{wang2022pico}, CR-DPLL \cite{wu2022revisiting}, PiCO+ \cite{wang2022pico+}, IRNet \cite{lian2022arnet}, ALIM-Scale and ALIM-Onehot \cite{xu2024alim}.

On five of the nine subtasks, our method achieves the best performances, while on the remaining subtasks, ALIM-Onehot or ALIM-Scale achieves the best performances (See Table \ref{synthetic}). It is worth noting that our method is not designed for synthetic datasets, but still achieves good performance. It can be clearly seen that our method has more advantages when $q$ is small. This is because there are usually relatively few candidate labels associated with each sample on the dataset annotated by the pre-trained model.

\subsection{Ablations}
We conduct experiments on four degenerations of our method to demonstrate the effectiveness of our proposed modules, which are: 1. w/o Co-PL: replaces the collaborative pseudo-labeling mechanism to performing pseudo-labeling with their own prediction; 2. w/ SupCont: replace noisy supervised contrastive learning to traditional supervised contrastive learning; 3. w/o proto: does not perform prototypical similarity alignment; 4. w/o U: discarding unlabeled set $\mathcal{U}$ during training.

As shown in Table \ref{tab:ablation_experiment}, our complete method (Co-Reg) consistently achieves the highest accuracy across all three pre-trained VLMs (CLIP ViT-B/32, CLIP ViT-B/16, and LLaVA-1.5) and all six datasets (CIFAR-10, CIFAR-100, SVHN, F-MNIST, EuroSAT, and GTSRB), which directly verifies the effectiveness of the proposed modules. For instance, under CLIP ViT-B/32, Co-Reg outperforms all degenerated versions: it reaches 71.04\% on CIFAR-100, which is 2.64\% higher than w/o Co-PL (68.40\%) and 5.72\% higher than w/o U (65.32\%); on SVHN, its accuracy (46.57\%) is 4.22\% higher than w/o Co-PL (42.35\%) and 1.13\% higher than w/o proto (45.44\%). A similar trend is observed in CLIP ViT-B/16: Co-Reg achieves 72.03\% on CIFAR-100 (surpassing w/o U by 5.73\%) and 50.74\% on GTSRB (outperforming w/o Co-PL by 3.58\%), while LLaVA-1.5-based Co-Reg leads by 2.79\% on CIFAR-100 (62.54\% vs. 59.75\% of w/o Co-PL) and 2.40\% on SVHN (86.70\% vs. 84.30\% of w/o U), confirming the robustness of our method across different VLMs.

The performance gaps between Co-Reg and each degenerated model further highlight the indispensable role of each proposed component. Removing the collaborative pseudo-labeling mechanism (w/o Co-PL) leads to noticeable accuracy drops on most datasets, indicating that collaborative pseudo-labeling effectively reduces the noise of individual predictions and improves label reliability. Replacing noisy supervised contrastive learning with traditional supervised contrastive learning (w/ SupCont) results in lower performance (e.g., 67.96\% vs. 71.04\% on CIFAR-100 under CLIP ViT-B/32), demonstrating the value of noise robustness in contrastive learning. Omitting prototypical similarity alignment (w/o proto) and discarding the unlabeled set (w/o U) also cause consistent performance degradation across VLMs—for example, w/o proto under LLaVA-1.5 achieves 61.50\% on CIFAR-100 (1.04\% lower than Co-Reg) and w/o U under CLIP ViT-B/16 reaches 45.90\% on GTSRB (4.84\% lower than Co-Reg)—which proves that both prototypical alignment and unlabeled data contribute to enhancing feature discrimination and model generalization.

\begin{figure*}[t]
    \centering
    \subfigure[$\lambda_u$ Sensitivity Analysis\label{subfig:lambda_u}]{
        \includegraphics[width=0.45\textwidth]{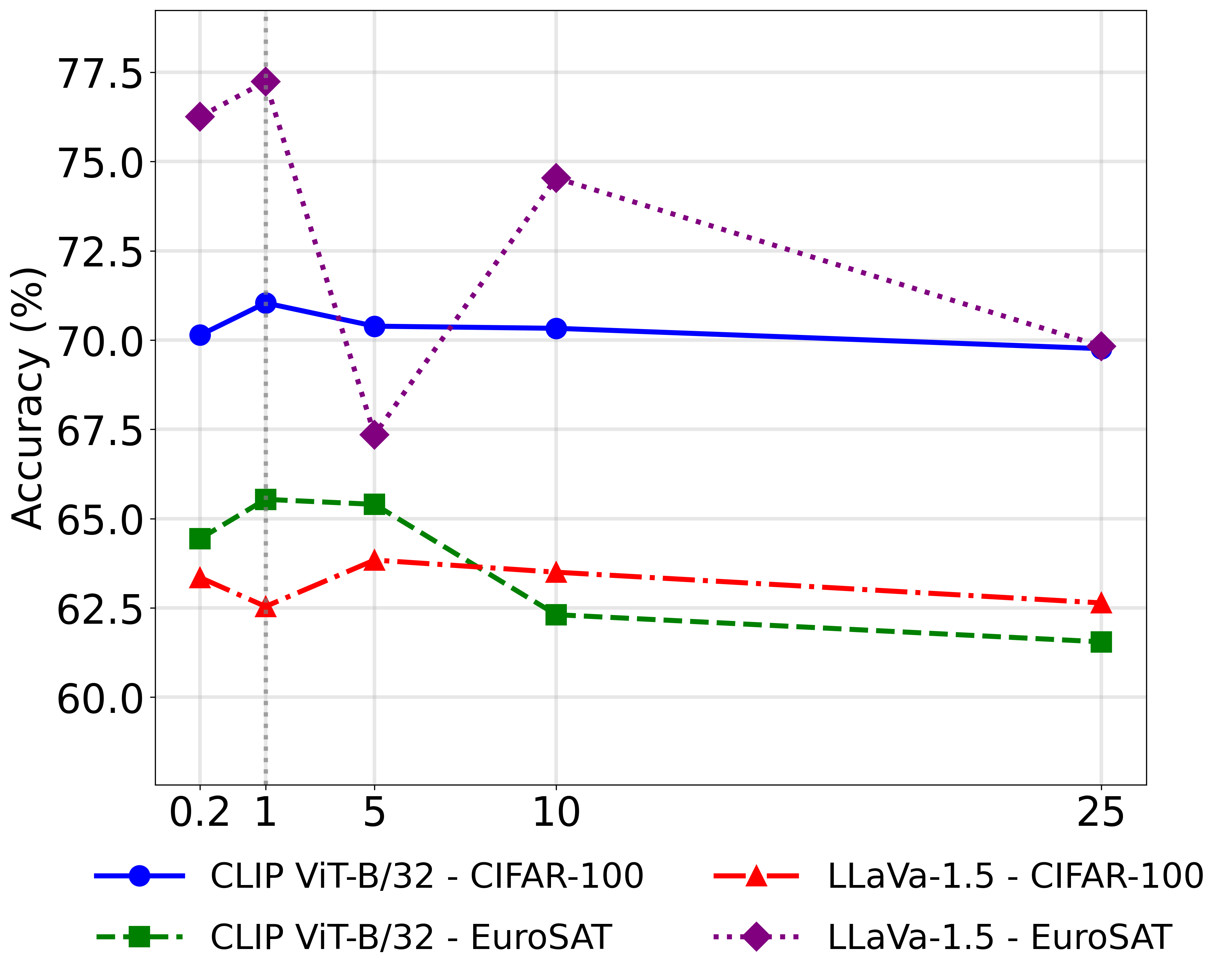}
    }
    \hfill
    \subfigure[$d'$ Sensitivity Analysis\label{subfig:low_dim}]{
        \includegraphics[width=0.45\textwidth]{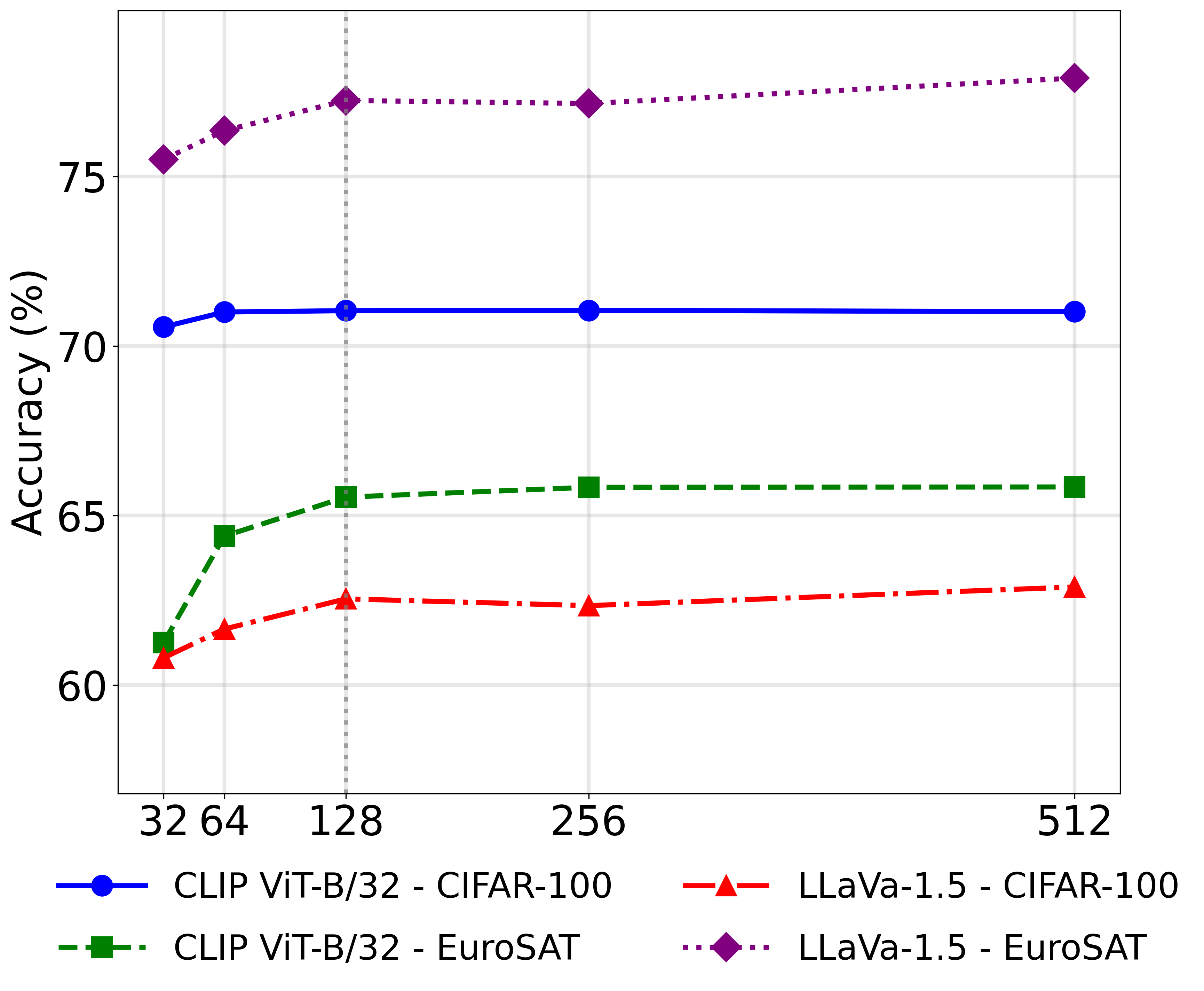}
    }
    \vspace{0.3cm}
    \subfigure[$\tau_{\text{div}}$ Sensitivity Analysis\label{subfig:p_threshold}]{
        \includegraphics[width=0.45\textwidth]{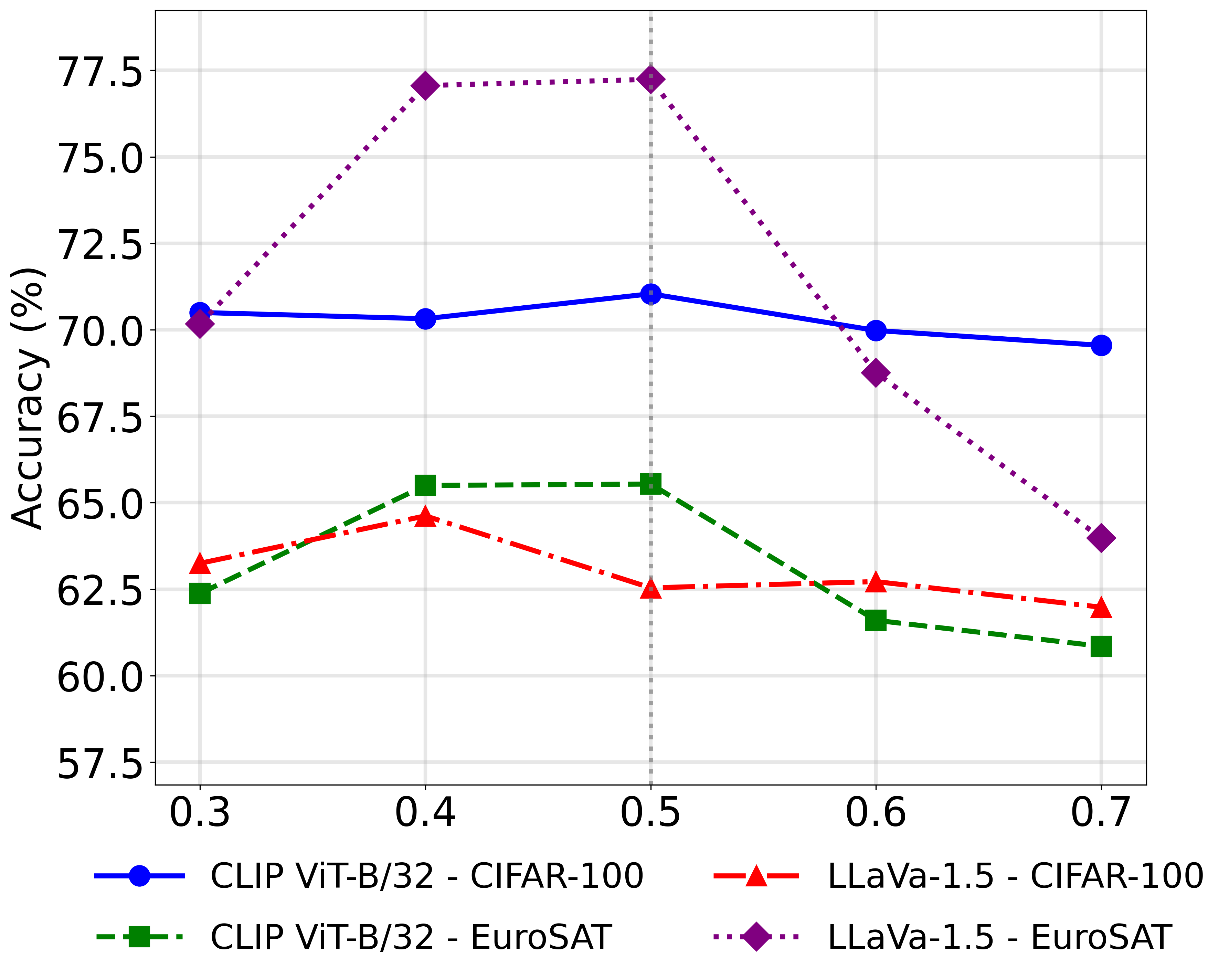}
    }
    \hfill
    \subfigure[$T$ Sensitivity Analysis\label{subfig:T}]{
        \includegraphics[width=0.45\textwidth]{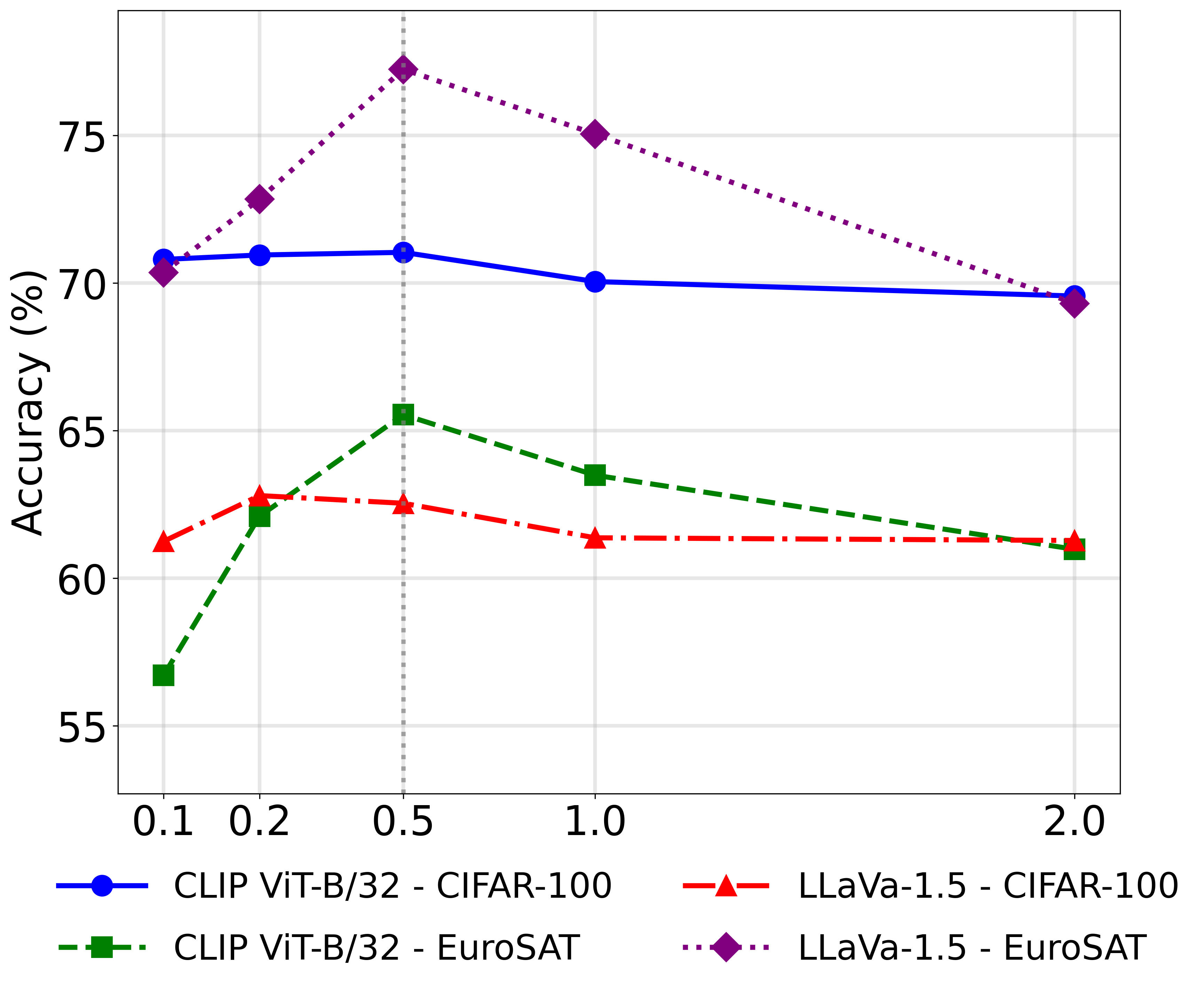}
    }
    \caption{Sensitivity analysis of key hyper-parameters in our proposed method. Each subfigure illustrates the performance variation with respect to specific hyper-parameters: $\lambda_u$ (weight coefficient of unlabeled set loss), $d'$ (feature dimension of projected representations for prototypical similarity alignment and noisy contrastive learning), $\tau_{\text{div}}$ (threshold for dividing partial and unlabeled set), and $T$ (sharpening temperature parameter in co-pseudo-labeling).}
    \label{fig:hyper_parameter_sensitivity}
\end{figure*}

\subsection{Hyper-parameter sensitivity}
To verify the robustness of our proposed method and rationalize the selection of key hyper-parameters, we conduct sensitivity analysis on four critical parameters: $\lambda_u$ (weight coefficient of unlabeled set loss), $d'$ (feature dimension of projected representations), $\tau_{\text{div}}$ (threshold for dividing partial and unlabeled sets), and $T$ (sharpening temperature in co-pseudo-labeling). Experiments are performed on two representative VLMs (CLIP ViT-B/32 and LLaVA-1.5) and two datasets (CIFAR-100 and EuroSAT), with results presented in Fig. \ref{fig:hyper_parameter_sensitivity} and detailed numerical values provided in the accompanying tables.

For $\lambda_u$, which balances the contribution of labeled and unlabeled data losses, Fig. \ref{subfig:lambda_u} and corresponding numerical results show that the default value ($\lambda_u=1$) achieves the optimal performance across most settings: it reaches 71.04\% on CLIP ViT-B/32 - CIFAR-100 and 77.24\% on LLaVA-1.5 - EuroSAT, outperforming other values. When $\lambda_u$ is set to 0.2 (underweighting unlabeled data), the accuracy drops to 70.14\% (CLIP ViT-B/32 - CIFAR-100) and 76.26\% (LLaVA-1.5 - EuroSAT), indicating insufficient utilization of unlabeled data. Conversely, excessively large $\lambda_u$ (e.g., 25) leads to more significant performance degradation (69.76\% on CLIP ViT-B/32 - CIFAR-100 and 69.83\% on LLaVA-1.5 - EuroSAT), as the noise introduced by unlabeled data overwhelms the supervised signal. This confirms that the default $\lambda_u=1$ effectively balances labeled supervision and unlabeled data augmentation.

Regarding $d'$ (projected feature dimension for prototypical alignment and noisy contrastive learning), Fig. \ref{subfig:low_dim} and numerical results demonstrate a gradual performance improvement as $d'$ increases from 32 to 128, followed by marginal fluctuations for larger values. The default $d'=128$ achieves 71.04\% (CLIP ViT-B/32 - CIFAR-100) and 77.24\% (LLaVA-1.5 - EuroSAT), which are comparable to the performance at $d'=256$ (71.05\% and 77.15\%) and $d'=512$ (71.01\% and 77.90\%). When $d'$ is too small (32), the feature representation capacity is limited, resulting in lower accuracy (70.56\% on CLIP ViT-B/32 - CIFAR-100 and 75.50\% on LLaVA-1.5 - EuroSAT). These results indicate that our method is not highly sensitive to $d'$ within a reasonable range (128–512), and the default $d'=128$ strikes a balance between representation capacity and computational efficiency.

For $\tau_{\text{div}}$, the threshold determining the division between partially labeled and unlabeled data, Fig. \ref{subfig:p_threshold} and numerical data show that the default value ($\tau_{\text{div}}=0.5$) delivers the best overall performance: 71.04\% on CLIP ViT-B/32 - CIFAR-100 and 77.24\% on LLaVA-1.5 - EuroSAT. Deviating from this threshold in either direction leads to performance declines: when $\tau_{\text{div}}=0.3$ (overly strict division), the accuracy drops to 70.50\% (CLIP ViT-B/32 - CIFAR-100) and 70.17\% (LLaVA-1.5 - EuroSAT) due to insufficient partially labeled data; when $\tau_{\text{div}}=0.7$ (overly loose division), the accuracy further decreases to 69.55\% and 63.98\% respectively, as low-confidence samples are incorrectly included in the partially labeled set, introducing noise. This validates that $\tau_{\text{div}}=0.5$ optimally balances the quantity and quality of partially labeled data.

Finally, for $T$ (temperature parameter controlling the sharpness of pseudo-labels in co-pseudo-labeling), Fig. \ref{subfig:T} and numerical results reveal that the default $T=0.5$ achieves the highest accuracy across all tested settings: 71.04\% (CLIP ViT-B/32 - CIFAR-100) and 77.24\% (LLaVA-1.5 - EuroSAT). When $T$ is too small (0.1), the pseudo-labels become overly sharp and overconfident, leading to performance degradation (56.71\% on CLIP ViT-B/32 - EuroSAT and 70.36\% on LLaVA-1.5 - EuroSAT). When $T$ is too large (2.0), the pseudo-labels become overly smooth and ambiguous, resulting in reduced accuracy (60.98\% on CLIP ViT-B/32 - EuroSAT and 69.31\% on LLaVA-1.5 - EuroSAT). This confirms that $T=0.5$ effectively balances the confidence and diversity of pseudo-labels, facilitating reliable collaborative pseudo-labeling.

Overall, the sensitivity analysis demonstrates that our proposed method maintains stable performance across a wide range of hyper-parameter values, and the selected default values are validated to be optimal for balancing performance and generalization. This robustness underscores the practicality and reliability of our method in real-world scenarios.

\begin{table*}[]
\centering
\setlength{\tabcolsep}{3pt}
\begin{tabular}{@{}p{0.15\textwidth}*{4}{>{\centering\arraybackslash}p{0.2\textwidth}}@{}}
\toprule
Paradigms               & Samples      & Human Annotations & Inference Size   & Perf. Improvements \\ \midrule
Zero-Shot               & $\times$     & $\times$          & -                & -                  \\
Few-shot FT & few          & few               & increase sightly & $\checkmark$       \\
KD$_{unsup}$       & $\checkmark$ & $\times$          & small            & $\times$           \\
KD$_{sup}$         & $\checkmark$ & $\checkmark$      & small            & $\checkmark$       \\
Fully FT       & $\checkmark$ & $\checkmark$      & -                & $\checkmark$       \\
NPLL                   & $\checkmark$ & $\times$          & small            & $\checkmark$       \\ \bottomrule
\end{tabular}
\caption{Comparison among different pre-trained model application paradigms. Zero-Shot indicates directly performing zero-shot inference on untrained tasks. Few-shot fine-tuning (FT) indicates techniques including prompt learning, adapter and LoRA. KD$_{sup}$ and KD$_{unsup}$ represent supervised and unsupervised knowledge distillation, i.e. with or without task-relevant labels, respectively. Fully fine-tuning (FT) indicates fine-tuning the whole model with all labeled samples of downstream tasks. "-" means remaining the same with original model.}
\label{paradigms}
\end{table*}

\section{Discussion and Limitation}
In this section, we briefly discuss the advantages and disadvantages of incorporating NPLL into distillation from pre-trained VLMs and compare this approach with other mainstream paradigms of applying pre-trained models to downstream tasks.

\subsection{Advantages of incorporating NPLL}
Just like what this paper does, we can use pre-trained models as weak annotators to annotate unlabeled samples of downstream task with candidate label sets, and then formalize this task as a NPLL problem and design corresponding algorithm to address it.

Table \ref{paradigms} compares different pre-trained model application paradigms. We can see that incorporating NPLL is the only one that can achieve performance improvements over the original model without using additional manual annotations. Meanwhile, by retraining specialized small models on the downstream samples, the inference model size are significantly reduced. Additionally, due to the fact that few-shot fine-tuning techniques (e.g., prompt learning, adaptors and LoRA) only attach a small number of trainable parameters to the pre-trained models, their performance improvements are usually limited.

It is worth noting that the main difference from traditional unsupervised KD is that this approach formalizes the downstream task as a specific weakly-supervised learning problem, i.e. NPLL, and employs elaborately designed consistency regularization methods, which can achieve significantly better performances on many scenarios compared to the original VLM. In contrast, KD uses the output class distributions of the pre-trained VLM as the training target, aiming to transfer the knowledge from a large model to a smaller specialized model and often does not achieve performance improvements.

\begin{figure*}[t]
    \centering
    \renewcommand{\thesubfigure}{}

    \setlength{\subfigcapskip}{-5pt}
    \setlength{\subfigtopskip}{0pt}
    \setlength{\subfigbottomskip}{0pt}
    
    \subfigure[]{\includegraphics[width=0.25\textwidth]{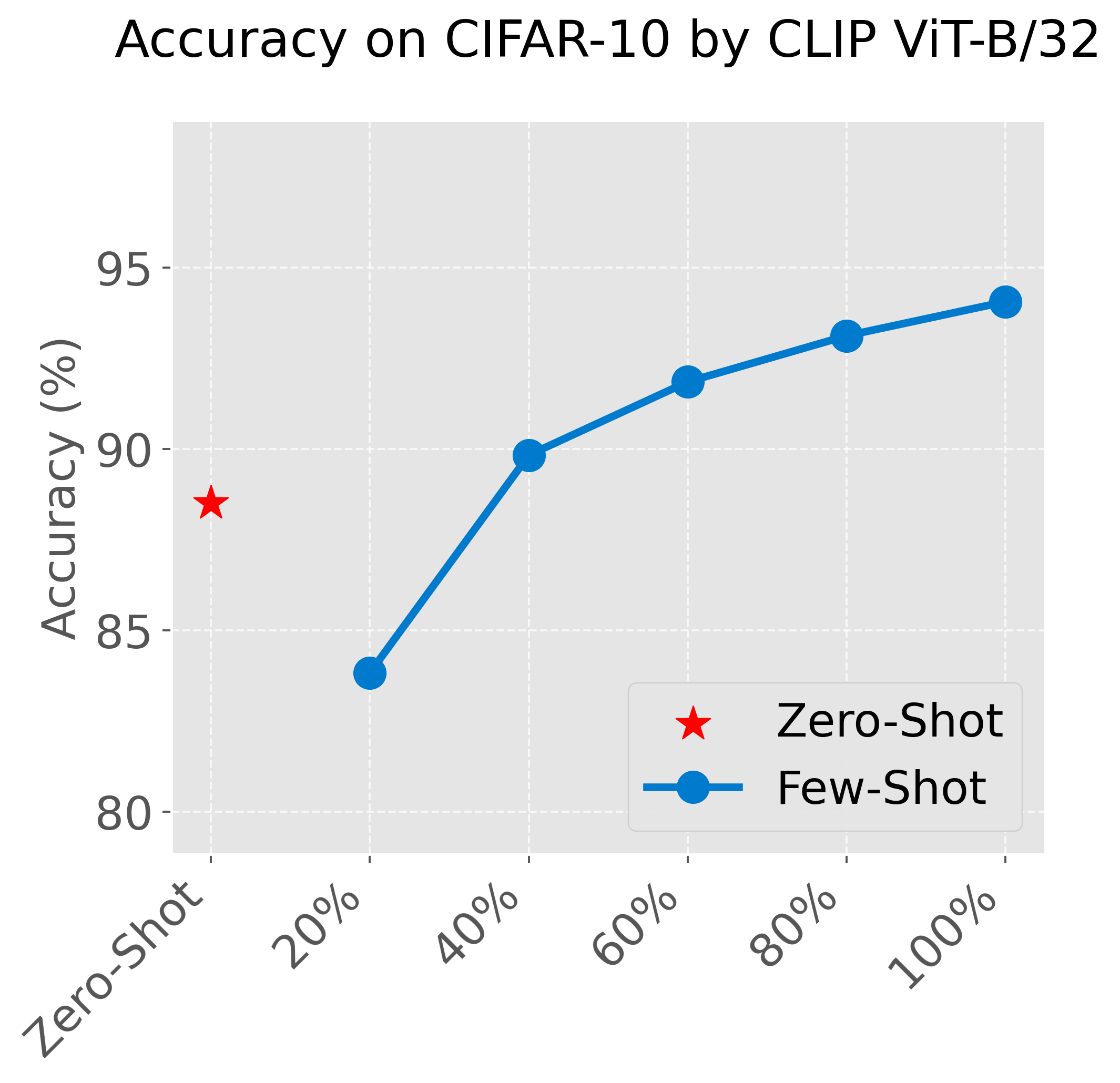}}%
    \hfill
    \subfigure[]{\includegraphics[width=0.25\textwidth]{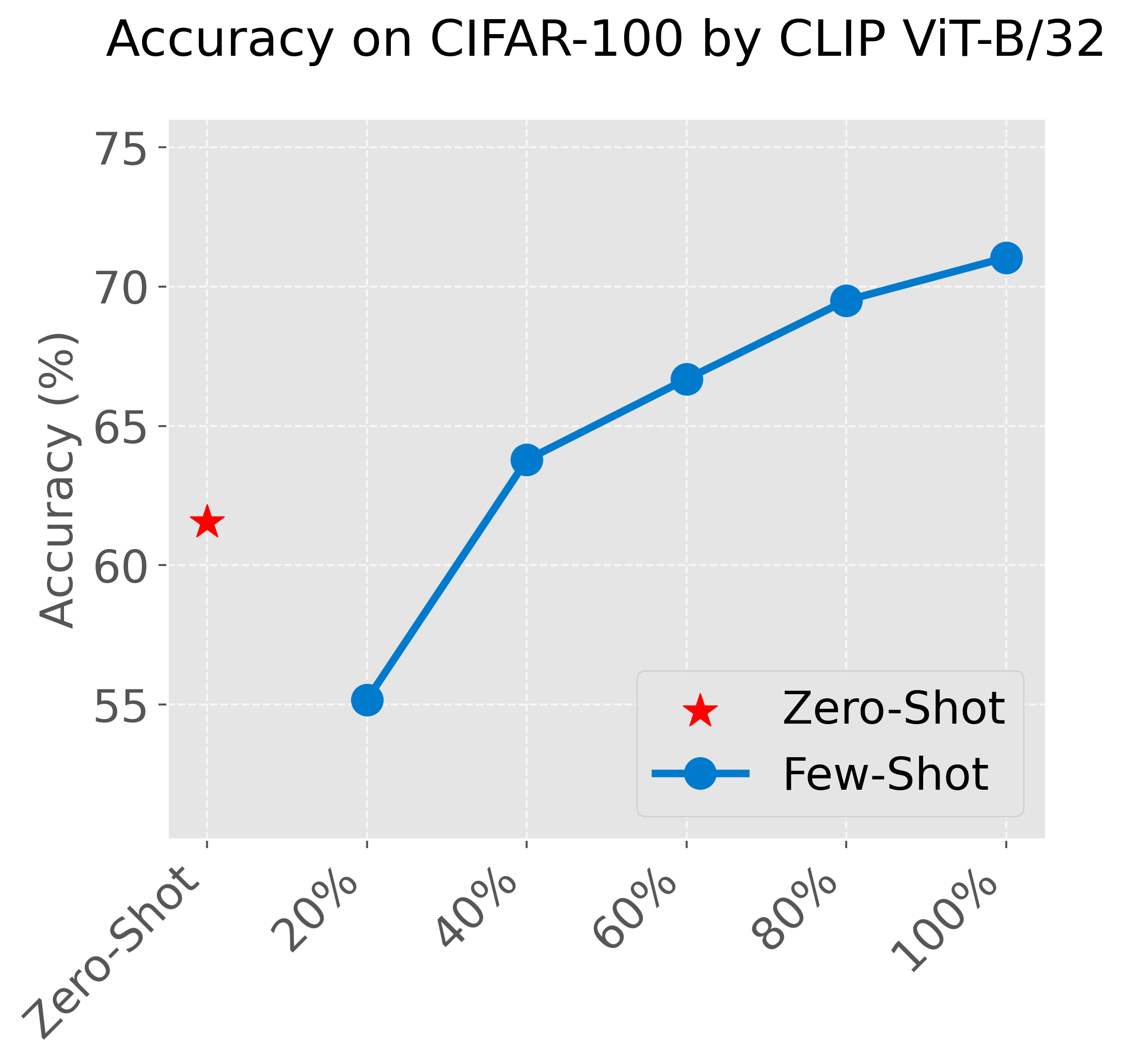}}%
    \hfill
    \subfigure[]{\includegraphics[width=0.25\textwidth]{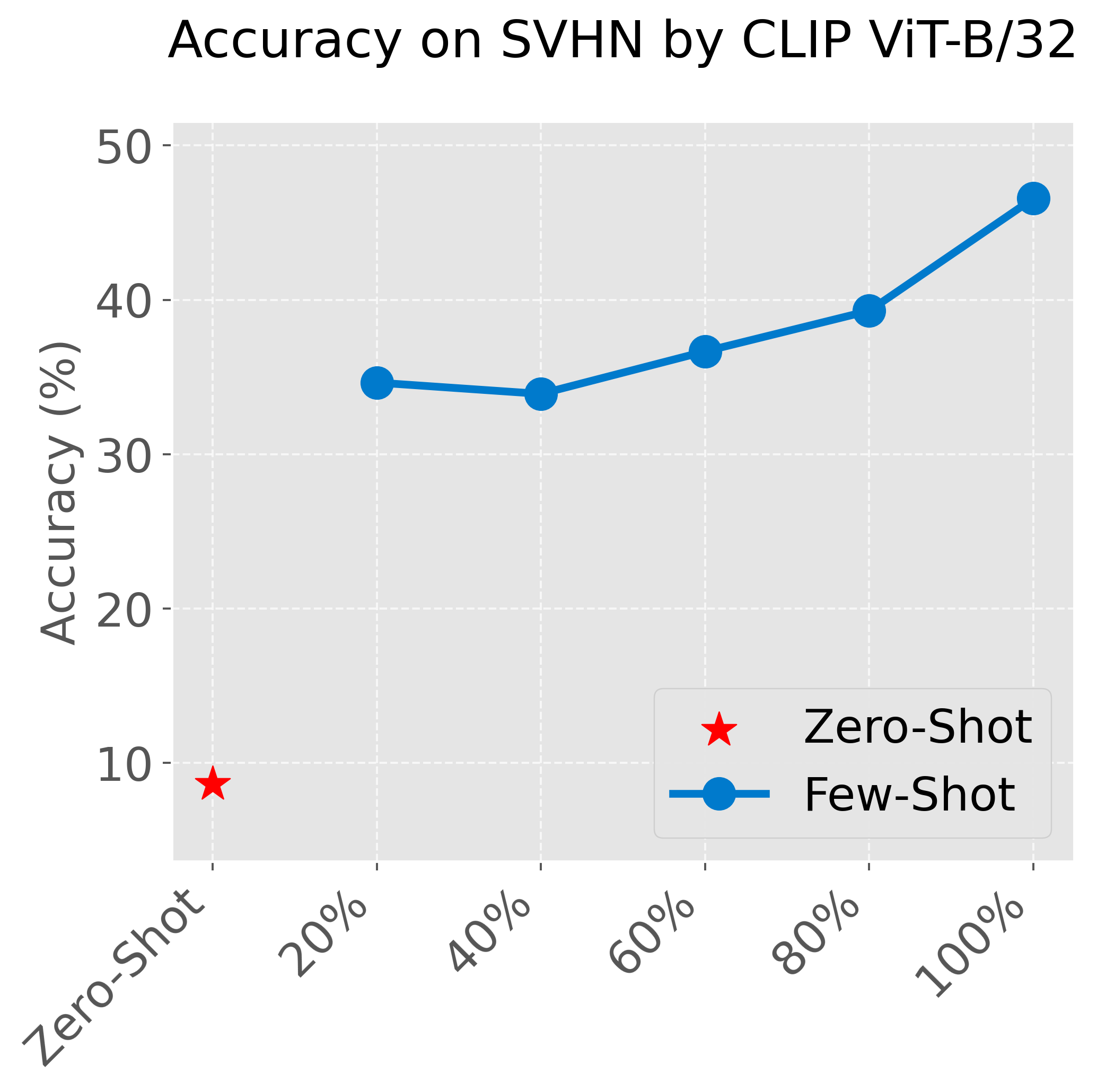}}%
    \hfill
    \subfigure[]{\includegraphics[width=0.25\textwidth]{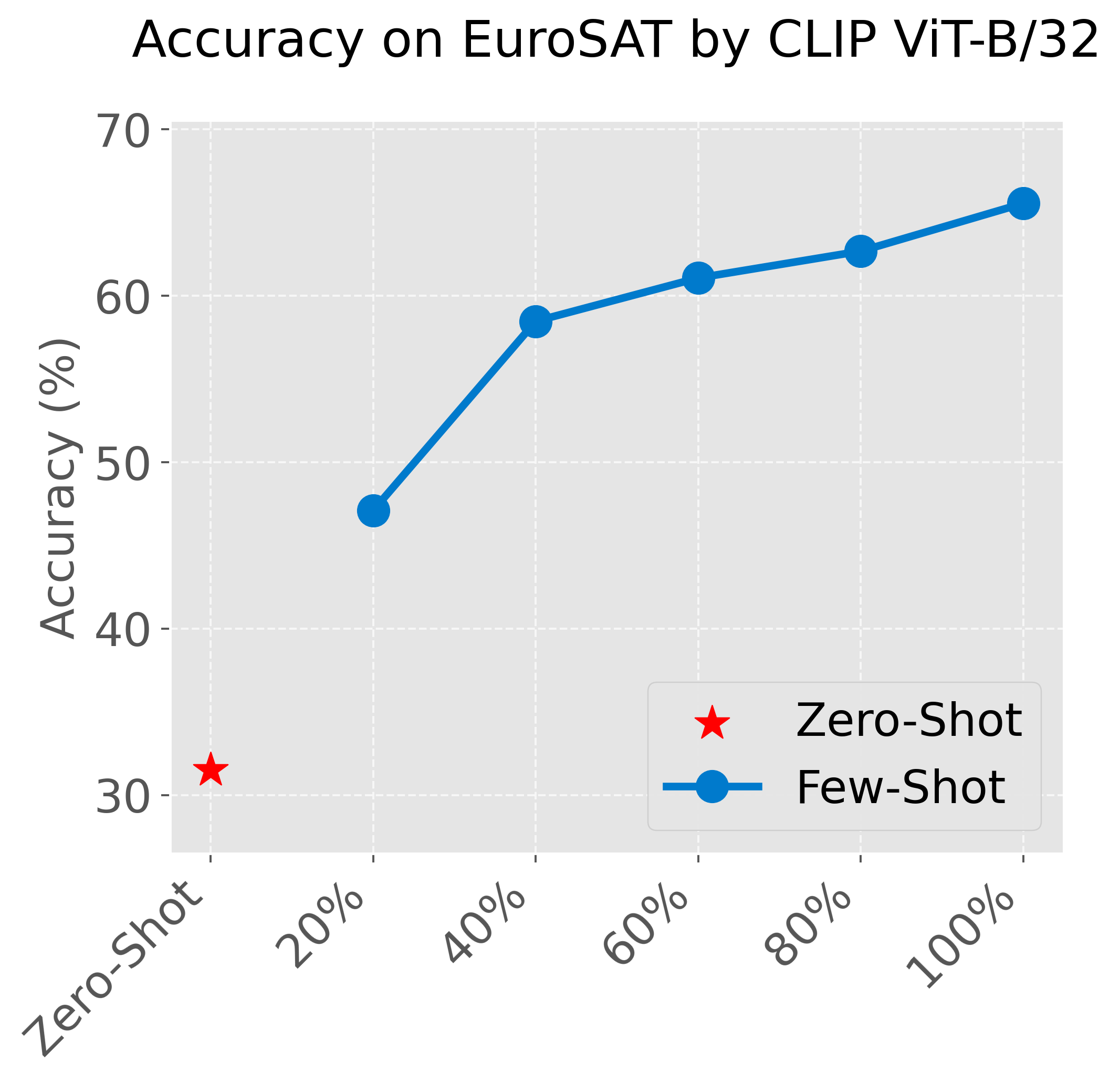}}
    
    %\text{CLIP ViT-B/32 Results} 
    
    \subfigure[]{\includegraphics[width=0.25\textwidth]{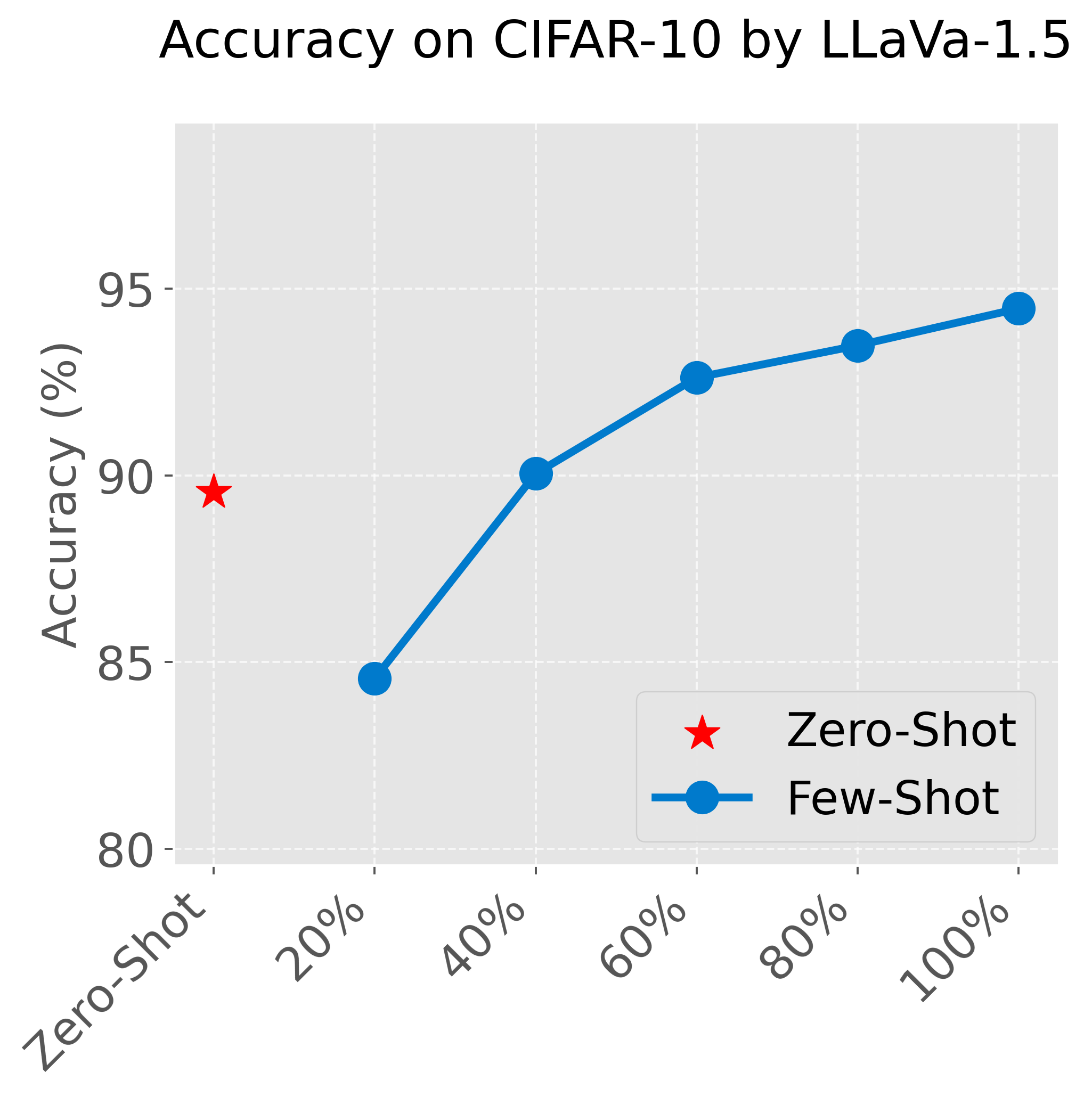}}%
    \hfill
    \subfigure[]{\includegraphics[width=0.25\textwidth]{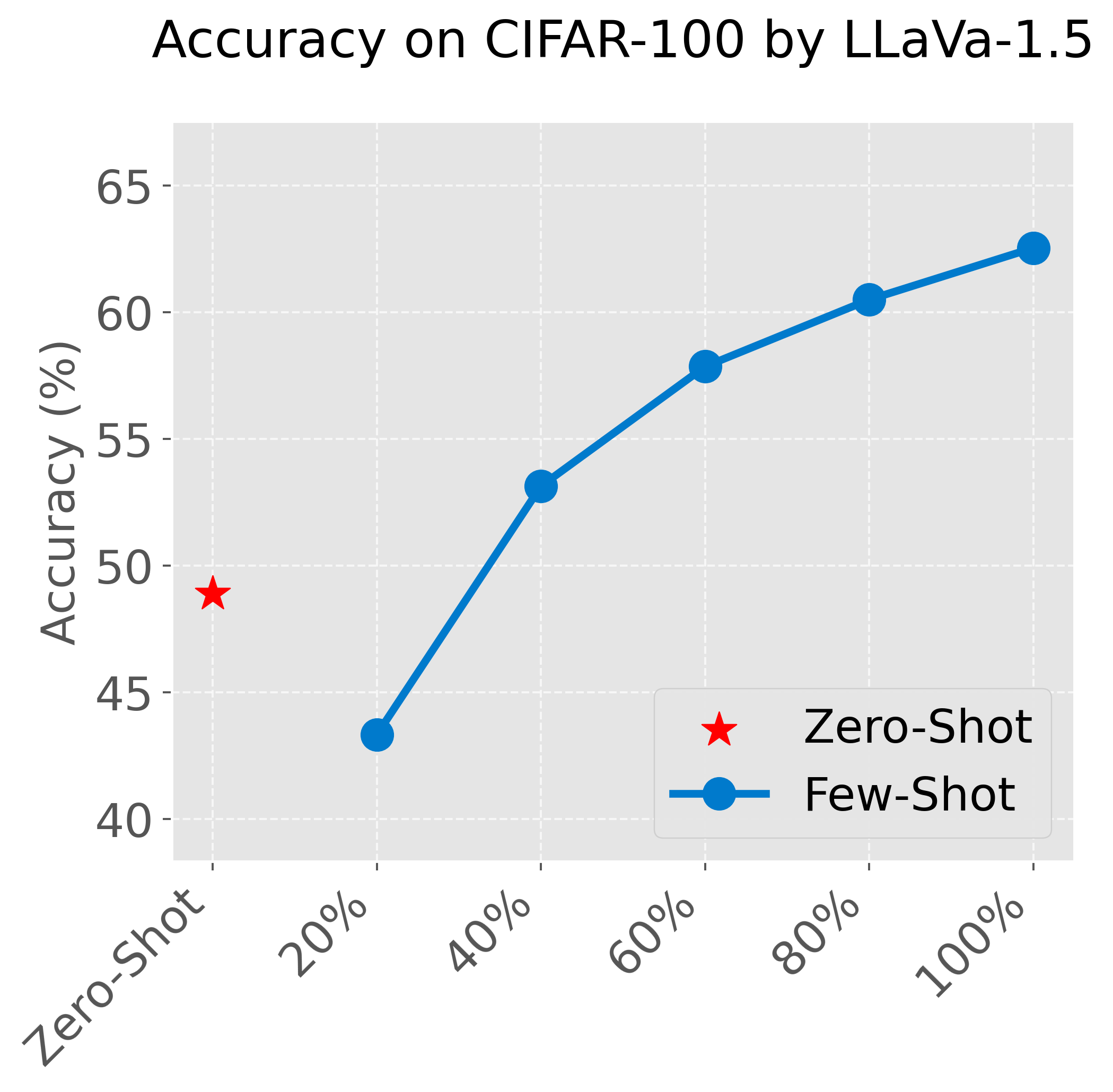}}%
    \hfill
    \subfigure[]{\includegraphics[width=0.25\textwidth]{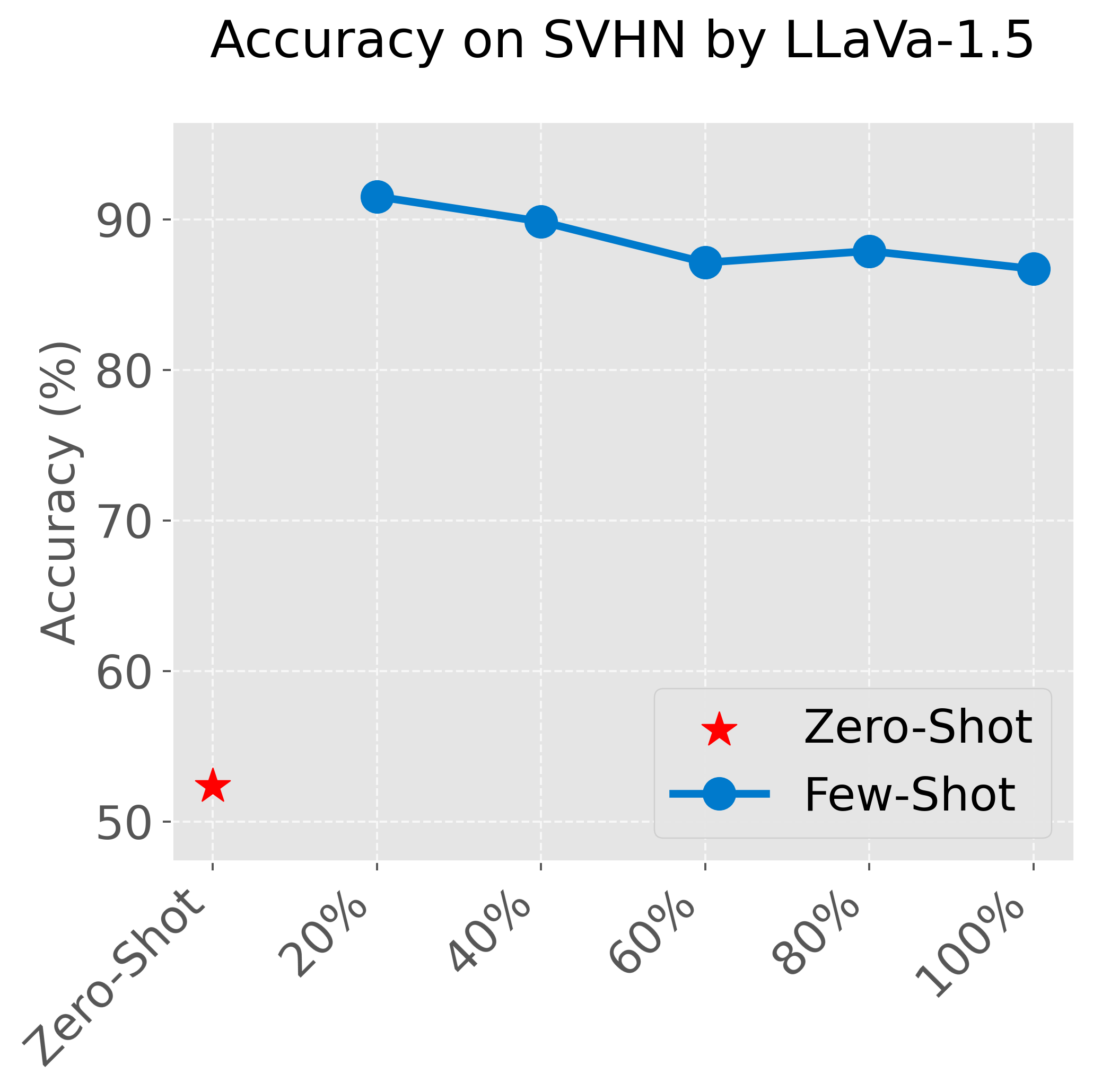}}%
    \hfill
    \subfigure[]{\includegraphics[width=0.25\textwidth]{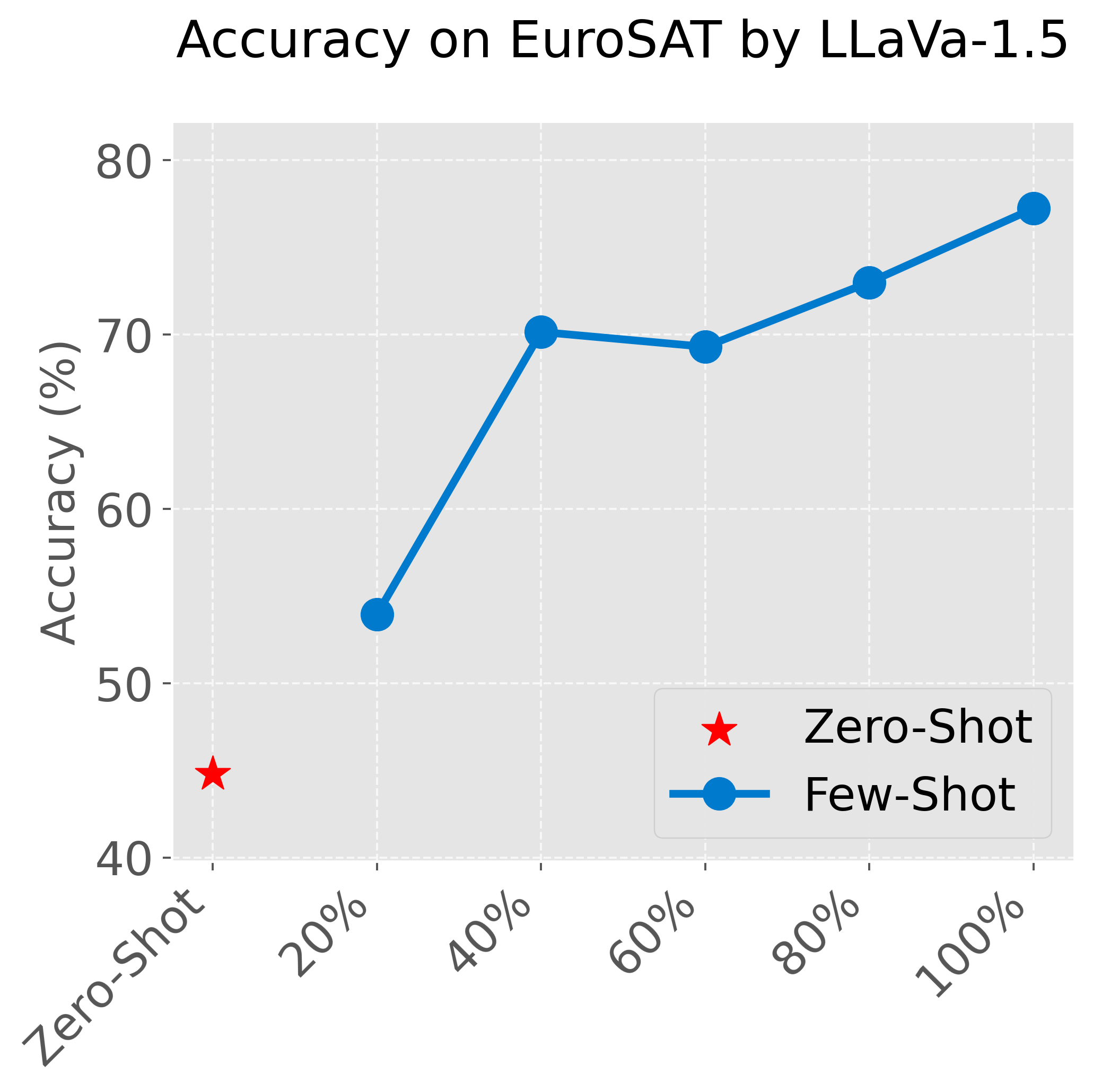}}
    
    %\text{LLaVa-1.5 Results} 
    
    \caption{Accuracy changes of our algorithm when using limited sample ratios (Zero-Shot, 20\%, 40\%, 60\%, 80\%, and 100\%) on the CIFAR-10, CIFAR-100, SVHN, and EuroSAT datasets. The red asterisk denotes Zero-Shot performance, while the blue line shows results for increasing sample ratios (20–100\%). Each subfigure corresponds to a model (CLIP ViT-B/32 or LLaVA-1.5) and dataset, illustrating accuracy improvements with growing labeled data.}
    \label{fig:data-limit-accuracy}
\end{figure*}

\subsection{Limitation}
Nevertheless, there are two main limitations associated with incorporating NPLL. First, in downstream tasks where the training images are relatively similar to the general domain images used for pre-training the large model, such as ImageNet, specialized models trained through NPLL often fail to surpass the performances of the original pre-trained model. In these kind of tasks, the large model can already yields satisfactory results via directly performing zero-shot inference. Our method should primarily be applied to tasks where the image domain significantly differs from the general domain, and where the pre-trained model does not perform well.

Second, we highlight a key limitation of our method: its reliance on a large number of downstream unlabeled samples to achieve competitive performance. As illustrated in Fig.\ref{fig:data-limit-accuracy}, on CIFAR-10 and CIFAR-100, our method requires at least 40\% of labeled samples to surpass the zero-shot performance of CLIP ViT-B/32 or LLaVA-1.5. For example, on CIFAR-100, our method achieves only 55.17\% accuracy with 20\% data (below CLIP’s zero-shot of 61.55\%), but rises to 63.78\% at 40\%, demonstrating a clear threshold for data sufficiency. Performance continues to improve monotonically with higher ratios, reaching 71.04\% by using all unlabeled training samples, but the 40\% baseline highlights the necessity of moderate data availability.

On EuroSAT, while our method exceeds zero-shot performance even with 20\% data using annotations from LLaVA (53.94\% vs. LLaVA’s 44.78\%), achieving satisfactory accuracy requires at least 40\% data. This suggests that although low ratios can surpass zero-shot baselines, meaningful performance gains still depend on accumulating more unlabeled samples for complex, context-rich datasets.

Notably, SVHN (simpler street number recognition) represents an exception: our method achieves 91.51\% accuracy with just 20\% data, far exceeding LLaVA’s Zero-Shot (52.35\%) and plateauing early. This indicates that for highly structured or low-variability tasks, our method can mitigate data limitations effectively, but for general visual recognition tasks (e.g., CIFAR, EuroSAT), a non-trivial number of unlabeled samples remains essential.

\section{Conclusion}
This paper proposes Co-Reg, a collaborative consistency regularization method for NPLL using annotations from pre-trained VLMs. By training two networks to collaboratively purify instance-dependent noisy labels via pseudo-labeling and enforcing consistency in label/feature spaces with class prototypes and contrastive learning, our method mitigates pre-trained model bias and optimizes downstream task representations. Experiments across noisy labeling manners and pre-trained models show our method outperforms state-of-the-art methods, especially when integrating few manual labels. This work bridges weakly-supervised learning and pre-trained model distillation, enabling efficient "annotation-free" training. Our approach not only advances NPLL but also provides inspiration for weakly-supervised learning research in the era of large models, highlighting new possibilities for leveraging pre-trained knowledge in weakly-supervised scenarios.

% \section*{Acknowledgments}
% This section should come before the References. Funding information may also be included here.

\section*{ORCID}
\noindent Qian-Wei Wang - \url{https://orcid.org/0009-0007-1760-0407}

\noindent Yaguang Song - \url{https://orcid.org/0000-0002-9300-8110}

\noindent Shu-Tao Xia - \url{https://orcid.org/0000-0002-8639-982X}

\bibliographystyle{ws-ijprai}
\bibliography{sample}

\appendix

\section{Expertimental Datasets and Pre-Trained VLMs}
\label{appendix-datasets}
To validate the necessity of each proposed module, six benchmark datasets are employed, covering diverse image domains and difficulty levels:

\begin{itemize}
    \item \textbf{CIFAR-10 \& CIFAR-100}: Standard 32×32 color image datasets proposed by the University of Toronto in 2009 , consisting of 60,000 samples each (50,000 for training and 10,000 for testing). CIFAR-10 covers 10 mutually exclusive classes (e.g., airplane, cat, truck), presenting a cross-species semantic classification task with large inter-class variance . CIFAR-100 further expands to 100 subclasses grouped into 20 superclasses (e.g., "reptiles" including lizards and snakes), featuring fine-grained categories with high intra-class similarity and providing both coarse and fine-grained labels . As classic benchmarks addressing MNIST's limitations (grayscale, low diversity), they are widely used for evaluating small-scale visual classification, dataset distillation, and neural architecture generalization.
    
    \item \textbf{SVHN}: A real-world digit classification dataset constructed from Google Street View house numbers, containing 10 digit classes (0–9) with 73,257 training samples, 26,032 test samples. Distinguished from synthetic datasets like MNIST, it exhibits natural variations in lighting, perspective, and background clutter.
    
    \item \textbf{F-MNIST}: A 28×28 grayscale fashion dataset released by Zalando as a modern alternative to MNIST , comprising 70,000 samples (60,000 training, 10,000 testing) across 10 apparel categories (e.g., T-shirts, sneakers, handbags). Each category contains 6,000 training and 1,000 test images, maintaining MNIST's structural format while presenting more complex texture and shape features . It is widely used for benchmarking image classification algorithms (e.g., CNNs, SVMs) and evaluating models' ability to capture fine-grained texture patterns, with applications in fashion product recognition and model interpretability research.
    
    \item \textbf{EuroSAT}: A remote sensing image dataset derived from Sentinel-2 satellite imagery , focusing on 10 land-use/land-cover classes (e.g., forests, highways, residential areas, pastures). It typically includes 27,000 64×64 RGB images, requiring models to capture long-range spatial correlations and spectral-spatial features inherent in satellite data. 
    
    \item \textbf{GTSRB}: The German Traffic Sign Recognition Benchmark, a critical dataset for traffic sign classification with 43 classes (e.g., speed limits, yield signs) . It contains 34,799 training samples, 4,410 validation samples, and 12,630 test samples, all resized to 32×32 RGB images. The dataset exhibits significant real-world variations: uneven class distribution, varying lighting conditions, viewpoint shifts, low-resolution artifacts, and physical damage (e.g., graffiti) . It emphasizes fine-grained visual pattern discrimination and is widely used to evaluate models for autonomous driving and intelligent transportation systems, often requiring data augmentation to address class imbalance.
\end{itemize}

Three representative Vision-Language Models (VLMs) are selected as backbones to verify cross-model generalization:

\begin{itemize}
    \item \textbf{CLIP ViT-B/32 \& CLIP ViT-B/16}: Foundational contrastively pre-trained VLMs that pioneer the paradigm of learning visual-text alignment from natural language supervision . Each model consists of two parallel encoders: a ViT-Base visual encoder (differentiated by patch sizes of 32×32 and 16×16) and a Transformer-based text encoder, jointly optimized to maximize cosine similarity of matched image-text pairs while minimizing that of mismatched pairs . Trained on 400 million image-text pairs from the WebImageText (WIT) dataset—spanning diverse visual concepts and textual descriptions—they exhibit strong zero-shot generalization to novel classification tasks without task-specific fine-tuning . The patch size disparity introduces a critical trade-off: ViT-B/16 captures finer-grained spatial details (beneficial for texture-rich targets) but incurs higher computational costs, while ViT-B/32 balances efficiency and semantic representation, making this pair ideal for exploring how feature granularity impacts downstream task adaptation.
    
    \item \textbf{LLaVA-1.5}: This instruction-tuned multimodal model represents an advanced VL alignment architecture . It integrates three core components: a high-resolution CLIP ViT-L/14 visual encoder (upgraded to 336px input resolution for enhanced detail perception), a 13B-parameter Vicuna v1.5 language model (fine-tuned from LLaMA-2), and a two-layer MLP cross-modal connector that replaces the original linear projection—this connector effectively maps visual features to the language model’s embedding space, resolving modality gap issues . Trained on ~650k diverse visual-language instruction samples (encompassing VQA, OCR, visual dialogue, and region-level reasoning), it specializes in following natural language prompts to perform VL-aligned tasks . Unlike CLIP’s contrastive pre-training, LLaVA-1.5’s instruction tuning enables human-like interactive reasoning, making it a critical testbed for evaluating our method’s efficacy on prompt-driven multimodal tasks.
\end{itemize}

\section{Details of Comparing Weakly-Supervised Methods}
This section provides brief descriptions of the comparative methods adopted in our experiments, covering partial label learning (PLL) methods for clean or noisy candidate labels and a state-of-the-art noisy label learning method for single labels annotated by VLMs.
\begin{itemize}
    \item \textbf{CR-DPLL} \cite{wu2022revisiting}: A deep PLL method that leverages consistency regularization to address label ambiguity. It decouples labels into non-candidate and candidate sets, applying a modified negative log-likelihood loss to non-candidate labels (exploiting their certainty as non-ground-truth) and a consistency loss to candidate labels. The regularization aligns model outputs across data augmentations via a dynamically inferred conformal label distribution, with a gradually increasing balancing factor to avoid early-stage performance degradation. It supports both uniform and instance-dependent partial labels without additional assumptions.
    
    \item \textbf{ALIM-Onehot} and \textbf{ALIM-Scale} \cite{xu2024alim}: Two variants of the ALIM framework for noisy PLL (where ground-truth may not be in candidate sets). They introduce a coefficient $\lambda$ to balance the reliability of candidate labels and model outputs, generating adjusted candidate sets for pseudo-label normalization (Onehot: max-value binarization; Scale: power-transformed normalization). Optional components include adaptive $\lambda$ estimation (via dataset noise level) and mixup training to enhance noise robustness.
    
    \item \textbf{DivideMix} \cite{li2020dividemix}: A state-of-the-art noisy label learning method used with VLM-annotated single labels. It frames noisy label learning as semi-supervised learning, training two diverged networks that dynamically partition data into clean (labeled) and noisy (unlabeled) sets via Gaussian Mixture Models (GMM) on per-sample losses. It improves MixMatch with label co-refinement (for clean samples) and co-guessing (for noisy samples) to mitigate overfitting to label noise.
\end{itemize}

\section{Prompt Templates and Class Names}
The specific prompt templates and class names used in our experiments are shown in Table \ref{tab:vlm_prompts}, \ref{tab:clip_class_names} and \ref{tab:llava_class_names}. For experiments under CLIP annotation, the ``\textless \textgreater'' placeholder is replaced with the corresponding class names of datasets, e.g. ``a photo of a airplane''; for experiments under LLaVA annotation, the ``\textless \textgreater'' placeholder is replaced with dataset-specific descriptions (e.g., ``image" or ``number"), followed by all class names of the dataset separated by commas, e.g. ``Which of the following classes does this image belong to? airplane, automobile, bird, ..., truck.''.

\begin{table*}[h]
    \centering
    \renewcommand{\arraystretch}{2.0}
    \begin{tabular}{p{2cm} p{10cm}}
        \toprule
        \centering \textbf{VLM Type} & 
        \multicolumn{1}{c}{\textbf{Prompt Templates}} \\
        \midrule
        \centering CLIP & 
        \fontfamily{courier}\selectfont
        ``a photo of a \textless \textgreater''; ``a rendering of a \textless \textgreater''; ``a cropped photo of the \textless \textgreater''; ``the photo of a \textless \textgreater''; ``a photo of a clean \textless \textgreater''; ``a photo of a dirty \textless \textgreater''; ``a dark photo of the \textless \textgreater''; ``a photo of my \textless \textgreater''; ``a photo of the cool \textless \textgreater''; ``a close-up photo of a \textless \textgreater''; ``a bright photo of the \textless \textgreater''; ``a cropped photo of a \textless \textgreater''; ``a photo of the \textless \textgreater''; ``a good photo of the \textless \textgreater''; ``a photo of one \textless \textgreater''; ``a close-up photo of the \textless \textgreater''; ``a rendition of the \textless \textgreater''; ``a photo of the clean \textless \textgreater''; ``a rendition of a \textless \textgreater''; ``a photo of a nice \textless \textgreater''; ``a good photo of a \textless \textgreater''; ``a photo of the nice \textless \textgreater''; ``a photo of the small \textless \textgreater''; ``a photo of the weird \textless \textgreater''; ``a photo of the large \textless \textgreater''; ``a photo of a cool \textless \textgreater''; ``a photo of a small \textless \textgreater'' \\
        \midrule
        \centering LLaVA & 
        \fontfamily{courier}\selectfont
        ``Which of the following classes does this \textless \textgreater belong to?''; ``Which of the following categories does this \textless \textgreater fall into?''; ``Classify this \textless \textgreater into one of the classes.''; ``This \textless \textgreater belongs to one of the categories in the brackets. Please identify which specific category it belongs to.''; ``Which of the following category descriptions is the content in this \textless \textgreater most similar to? Please output the category name.''; ``Which of the following class names can best describe this \textless \textgreater? Please output the class name.''; ``This \textless \textgreater belongs to one of the following categories. Please identify which category this picture belongs to and output the category name. If it cannot be accurately classified into a specific category, please output several possible category names.''; ``You are labeling samples for an image classification task. Please accurately classify this \textless \textgreater into one of the following categories.'' \\
        \bottomrule
    \end{tabular}
    \caption{Prompt templates used for CLIP and LLaVA-based annotations. For experiments under CLIP annotation, the ``\textless \textgreater'' placeholder is replaced with the corresponding class names of datasets; for experiments under LLaVA annotation, the ``\textless \textgreater'' placeholder is replaced with dataset-specific descriptions (e.g., ``image'' or ``number''), followed by all class names of the dataset separated by commas.}
    \label{tab:vlm_prompts}
\end{table*}

\begin{table*}[h]
    \centering
    \small
    \renewcommand{\arraystretch}{1.8}
    \begin{tabular}{p{1cm} p{11cm}}
        \toprule
        \centering \textbf{Dataset} & \multicolumn{1}{c}{\textbf{Class Names}} \\ % 让表头居中
        \midrule
        \centering CIFAR-10 & 
        \fontfamily{courier}\selectfont
        ``airplane'', ``automobile'', ``bird'', ``cat'', ``deer'', ``dog'', ``frog'', ``horse'', ``ship'', ``truck'' \\
        \midrule
        \centering CIFAR-100 & 
        \fontfamily{courier}\selectfont
        ``apple'', ``aquarium\_fish'', ``baby'', ``bear'', ``beaver'', ``bed'', ``bee'', ``beetle'', ``bicycle'', ``bottle'', ``bowl'', ``boy'', ``bridge'', ``bus'', ``butterfly'', ``camel'', ``can'', ``castle'', ``caterpillar'', ``cattle'', ``chair'', ``chimpanzee'', ``clock'', ``cloud'', ``cockroach'', ``couch'', ``crab'', ``crocodile'', ``cup'', ``dinosaur'', ``dolphin'', ``elephant'', ``flatfish'', ``forest'', ``fox'', ``girl'', ``hamster'', ``house'', ``kangaroo'', ``keyboard'', ``lamp'', ``lawn\_mower'', ``leopard'', ``lion'', ``lizard'', ``lobster'', ``man'', ``maple\_tree'', ``motorcycle'', ``mountain'', ``mouse'', ``mushroom'', ``oak\_tree'', ``orange'', ``orchid'', ``otter'', ``palm\_tree'', ``pear'', ``pickup\_truck'', ``pine\_tree'', ``plain'', ``plate'', ``poppy'', ``porcupine'', ``possum'', ``rabbit'', ``raccoon'', ``ray'', ``road'', ``rocket'', ``rose'', ``sea'', ``seal'', ``shark'', ``shrew'', ``skunk'', ``skyscraper'', ``snail'', ``snake'', ``spider'', ``squirrel'', ``streetcar'', ``sunflower'', ``sweet\_pepper'', ``table'', ``tank'', ``telephone'', ``television'', ``tiger'', ``tractor'', ``train'', ``trout'', ``tulip'', ``turtle'', ``wardrobe'', ``whale'', ``willow\_tree'', ``wolf'', ``woman'', ``worm'' \\
        \midrule
        \centering SVHN & 
        \fontfamily{courier}\selectfont
        ``number 0'', ``number 1'', ``number 2'', ``number 3'', ``number 4'', ``number 5'', ``number 6'', ``number 7'', ``number 8'', ``number 9'' \\
        \midrule
        \centering F-MNIST & 
        \fontfamily{courier}\selectfont
        ``t-shirt/top'', ``trouser'', ``pullover'', ``dress'', ``coat'', ``sandal'', ``shirt'', ``sneaker'', ``bag'', ``ankle\_boot'' \\
        \midrule
        \centering EuroSAT & 
        \fontfamily{courier}\selectfont
        ``AnnualCrop'', ``Forest'', ``HerbaceousVegetation'', ``Highway'', ``Industrial'', ``Pasture'', ``PermanentCrop'', ``Residential'', ``River'', ``SeaLake'' \\
        \midrule
        \centering GTSRB & 
        \fontfamily{courier}\selectfont
        ``German Traffic Sign: Speed limit 20'', ``German Traffic Sign: Speed limit 30'', ``German Traffic Sign: Speed limit 50'', ``German Traffic Sign: Speed limit 60'', ``German Traffic Sign: Speed limit 70'', ``German Traffic Sign: Speed limit 80'', ``German Traffic Sign: Speed limit 80 cancelled'', ``German Traffic Sign: Speed limit 100'', ``German Traffic Sign: Speed limit 120'', ``German Traffic Sign: No passing'', ``German Traffic Sign: No truck overtaking'', ``German Traffic Sign: Only have priority at the next intersection'', ``German Traffic Sign: Priority road'', ``German Traffic Sign: Give way'', ``German Traffic Sign: Stop'', ``German Traffic Sign: All vehicles prohibited'', ``German Traffic Sign: Trucks prohibited'', ``German Traffic Sign: No entry'', ``German Traffic Sign: Caution!'', ``German Traffic Sign: Curve (to the left)'', ``German Traffic Sign: Curve (to the right)'', ``German Traffic Sign: Continuous curves'', ``German Traffic Sign: Rough road'', ``German Traffic Sign: Slippery road'', ``German Traffic Sign: The road narrows on the right'', ``German Traffic Sign: Construction site'', ``German Traffic Sign: Signal light'', ``German Traffic Sign: Pay attention to pedestrians'', ``German Traffic Sign: Pay attention to children'', ``German Traffic Sign: Watch out for bikes'', ``German Traffic Sign: Watch out for snow/ice on the road'', ``German Traffic Sign: Deer ahead'', ``German Traffic Sign: Unlimited speed'', ``German Traffic Sign: Turn right'', ``German Traffic Sign: Turn left'', ``German Traffic Sign: Go straight'', ``German Traffic Sign: Go straight and turn right'', ``German Traffic Sign: Go straight and turn left'', ``German Traffic Sign: Drive on the right'', ``German Traffic Sign: Drive on the left'', ``German Traffic Sign: Driving around the island'', ``German Traffic Sign: Lift ban on no passing'', ``German Traffic Sign: Lift ban on truck overtaking'' \\
        \bottomrule
    \end{tabular}
    \caption{Class names of benchmark datasets for CLIP-based annotations.}
    \label{tab:clip_class_names}
\end{table*}

\begin{table*}[h]
    \centering
    \small
    \renewcommand{\arraystretch}{1.8} % 调整行高适配内容
    \begin{tabular}{p{1cm} p{1cm} p{10cm}} % 增加了 Description 列
        \toprule
        \centering \textbf{Dataset} & \centering \textbf{Description} & \multicolumn{1}{c}{\textbf{Class Names}} \\ % 修改表头，增加 Description 列
        \midrule
        \centering CIFAR-10 & 
        \centering image & 
        \fontfamily{courier}\selectfont
        ``airplane'', ``automobile'', ``bird'', ``cat'', ``deer'', ``dog'', ``frog'', ``horse'', ``ship'', ``truck'' \\
        \midrule
        \centering CIFAR-100 & 
        \centering image & 
        \fontfamily{courier}\selectfont
        ``apple'', ``aquarium fish'', ``baby'', ``bear'', ``beaver'', ``bed'', ``bee'', ``beetle'', ``bicycle'', ``bottle'', ``bowl'', ``boy'', ``bridge'', ``bus'', ``butterfly'', ``camel'', ``can'', ``castle'', ``caterpillar'', ``cattle'', ``chair'', ``chimpanzee'', ``clock'', ``cloud'', ``cockroach'', ``couch'', ``crab'', ``crocodile'', ``cup'', ``dinosaur'', ``dolphin'', ``elephant'', ``flatfish'', ``forest'', ``fox'', ``girl'', ``hamster'', ``house'', ``kangaroo'', ``keyboard'', ``lamp'', ``lawn mower'', ``leopard'', ``lion'', ``lizard'', ``lobster'', ``man'', ``maple tree'', ``motorcycle'', ``mountain'', ``mouse'', ``mushroom'', ``oak tree'', ``orange'', ``orchid'', ``otter'', ``palm tree'', ``pear'', ``pickup truck'', ``pine tree'', ``plain'', ``plate'', ``poppy'', ``porcupine'', ``possum'', ``rabbit'', ``raccoon'', ``ray'', ``road'', ``rocket'', ``rose'', ``sea'', ``seal'', ``shark'', ``shrew'', ``skunk'', ``skyscraper'', ``snail'', ``snake'', ``spider'', ``squirrel'', ``streetcar'', ``sunflower'', ``sweet pepper'', ``table'', ``tank'', ``telephone'', ``television'', ``tiger'', ``tractor'', ``train'', ``trout'', ``tulip'', ``turtle'', ``wardrobe'', ``whale'', ``willow tree'', ``wolf'', ``woman'', ``worm'' \\
        \midrule
        \centering SVHN & 
        \centering number & 
        \fontfamily{courier}\selectfont
        ``0'', ``1'', ``2'', ``3'', ``4'', ``5'', ``6'', ``7'', ``8'', ``9'' \\
        \midrule
        \centering F-MNIST & 
        \centering image & 
        \fontfamily{courier}\selectfont
        ``t-shirt/top'', ``trouser'', ``pullover'', ``dress'', ``coat'', ``sandal'', ``shirt'', ``sneaker'', ``bag'', ``ankle boot'' \\
        \midrule
        \centering EuroSAT & 
        \centering image & 
        \fontfamily{courier}\selectfont
        ``annual crop'', ``forest'', ``herbaceous vegetation'', ``highway'', ``industrial buildings'', ``pasture'', ``permanent crop'', ``residential buildings'', ``river'', ``sea'' \\
        \midrule
        \centering GTSRB & 
        \centering image & 
        \fontfamily{courier}\selectfont
        ``Traffic Sign: Speed limit 20'', ``Traffic Sign: Speed limit 30'', ``Traffic Sign: Speed limit 50'', ``Traffic Sign: Speed limit 60'', ``Traffic Sign: Speed limit 70'', ``Traffic Sign: Speed limit 80'', ``Traffic Sign: Speed limit 80 cancelled'', ``Traffic Sign: Speed limit 100'', ``Traffic Sign: Speed limit 120'', ``Traffic Sign: No passing'', ``Traffic Sign: No truck overtaking'', ``Traffic Sign: Priority at the next intersection'', ``Traffic Sign: Priority road'', ``Traffic Sign: Give way'', ``Traffic Sign: Stop'', ``Traffic Sign: Vehicles prohibited'', ``Traffic Sign: Trucks prohibited'', ``Traffic Sign: No entry'', ``Traffic Sign: Caution!'', ``Traffic Sign: Curve to left'', ``Traffic Sign: Curve to right'', ``Traffic Sign: Continuous curves'', ``Traffic Sign: Rough road'', ``Traffic Sign: Slippery road'', ``Traffic Sign: Narrows on the right'', ``Traffic Sign: Construction site'', ``Traffic Sign: Signal light'', ``Traffic Sign: Pedestrians'', ``Traffic Sign: Children'', ``Traffic Sign: Watch out for bikes'', ``Traffic Sign: Watch out for snow and ice'', ``Traffic Sign: Deer ahead'', ``Traffic Sign: Unlimited speed'', ``Traffic Sign: Turn right'', ``Traffic Sign: Turn left'', ``Traffic Sign: Go straight'', ``Traffic Sign: Go straight and turn right'', ``Traffic Sign: Go straight and turn left'', ``Traffic Sign: Drive right'', ``Traffic Sign: Drive left'', ``Traffic Sign: Drive around'', ``Traffic Sign: Lift ban on no passing'', ``Traffic Sign: Lift ban on truck overtaking'' \\
        \bottomrule
    \end{tabular}
    \caption{Class names and descriptions of benchmark datasets for LLaVA-based annotations.}
    \label{tab:llava_class_names}
\end{table*}

\section{Implementation Details}
\subsection{Data augmentation}
We use RandAugment \cite{cubuk2020randaugment} in our strong data augmentation. Here we list the specific PIL transforms we use in Table \ref{tab:data_augmentation}.

\begin{table}[h]
    \centering
    \renewcommand{\arraystretch}{1.5} % 调整行高
    \begin{tabular}{|c|c|c|}
        \hline
        \textbf{Transformation} & \textbf{Min Value} & \textbf{Max Value} \\
        \hline
        \fontfamily{courier}\selectfont AutoContrast & 0 & 1 \\
        \hline
        \fontfamily{courier}\selectfont Brightness & 0.05 & 0.95 \\
        \hline
        \fontfamily{courier}\selectfont Color & 0.05 & 0.95 \\
        \hline
        \fontfamily{courier}\selectfont Contrast & 0.05 & 0.95 \\
        \hline
        \fontfamily{courier}\selectfont Equalize & 0 & 1 \\
        \hline
        \fontfamily{courier}\selectfont Identity & 0 & 1 \\
        \hline
        \fontfamily{courier}\selectfont Posterize & 4 & 8 \\
        \hline
        \fontfamily{courier}\selectfont Rotate & -30 & 30 \\
        \hline
        \fontfamily{courier}\selectfont Sharpness & 0.05 & 0.95 \\
        \hline
        \fontfamily{courier}\selectfont ShearX & -0.3 & 0.3 \\
        \hline
        \fontfamily{courier}\selectfont ShearY & -0.3 & 0.3 \\
        \hline
        \fontfamily{courier}\selectfont Solarize & 0 & 256 \\
        \hline
        \fontfamily{courier}\selectfont TranslateX & -0.3 & 0.3 \\
        \hline
        \fontfamily{courier}\selectfont TranslateY & -0.3 & 0.3 \\
        \hline
    \end{tabular}
    \caption{Data Augmentation Operations with Min and Max Values}
    \label{tab:data_augmentation}
\end{table}

\subsection{Hyper-parameters}
In this section, we provide a detailed list of all the hyperparameters used in our experiments.

\begin{itemize}
    \item \textbf{Backbone:} We use the \textbf{PreAct ResNet-18} \cite{he2016identity} as the backbone for all comparing methods.
    
    \item \textbf{Learning Rate:} The learning rate is set to $lr = 0.1$, with a cosine schedule for learning rate decay during training.
    
    \item \textbf{Batch Size:} The training batch size is set to 256.
    
    \item \textbf{Epochs:} The number of warm-up epochs is chosen from 20 to 100, and the total number of training epochs is selected from 100 to 800.
    
    \item \textbf{Threshold for Dividing Partial and Unlabeled Set:} The threshold is set as $\tau_{div} = 0.5$ to divide the partial and unlabeled subsets in co-pseudo-labeling.
    
    \item \textbf{Number of Weakly-Augmented Inputs:} The number of weakly-augmented inputs for averaging the predicted probabilites of co-pseudo-labeling is $K = 2$.
    
    \item \textbf{Sharpening Temperature for Pseudo-Labels:} The sharpening temperature for pseudo-labels of co-pseudo-labeling is $T = 0.5$, controlling the smoothness of generated pseudo-labels.
    
    \item \textbf{Weight Parameter for Unlabeled Set:} The weight parameter of samples in unlabeled set is set to $\lambda_u = 1$, balancing the contribution of unlabeled set.
    
    \item \textbf{Projected Representation Dimension:} The dimension of projected representations when performing prototypical similarity alignment and noisy contrastive learning is $d' = 128$.
    
    \item \textbf{Feature Representation Queue Length:} The length of the feature representation queue, updated by the momentum encoder, is 8192.
    
    \item \textbf{MixUp:} For the MixUp data augmentation method, the Beta distribution parameter is set to $\alpha = 4$, controlling the mixing ratio of the samples.
    
    \item \textbf{Loss Function Weights:}
    \begin{itemize}
        \item Prototypical similarity alignment weight: $\beta_1 = 0.1$
        \item Noisy contrastive learning weight: $\beta_2 = 0.1$
    \end{itemize}
    
    \item \textbf{Momentum for Prototypes:} The momentum parameter for computing the moving average of prototypes is $\gamma = 0.999$.
    
    \item \textbf{Sharpening Temperature for Prototypical Similarity:} The sharpening temperature for prototypical similarity is $\tau_{proto} = 0.3$.
    
    \item \textbf{Momentum Encoder:} The momentum for updating the momentum encoder in noisy contrastive learning is set to $0.999$.
\end{itemize}

% \vspace*{-0.1in}
% \noindent
% \rule{12.6cm}{.1mm}

% \section*{Biographical Sketch and Photo}

% Upon acceptance of an article, a brief biographical sketch and
% photograph of each author are to be supplied to the Publisher.

% \biophoto{wang}{{\bf Chuan-Cheng Wang} received the
% B.S.~degree\break
% in electrical engineering from National Sun Yat-Sen University,
% Kaoh-\break siung, Taiwan in 1992 and the M.S. degree in
% computer science from National Chiao Tung University, Hsinchu,\break
% Taiwan in 2001.}

% \vglue-1.75truein
% \hspace*{2.45truein}
% \biophoto{gao}{{\bf Yongsheng Gao} received the B.Sc. and\break
% M.Sc. degrees in electronic engineering from\break Zhejiang
% University,\break China, in 1985 and 1988 respectively, and the
% Ph.D. in computer engineering from Nanyang Technological University,
% Singapore. Currently, he is an assistant professor with Nanyang
% Technological University, Singapore.}

\end{document}